\documentclass[10pt,twocolumn,letterpaper]{article}
\PassOptionsToPackage{numbers, compress}{natbib}
\usepackage{iccv}      
\usepackage{times}

\usepackage{graphicx}
\usepackage{amsmath}
\usepackage{amssymb}
\usepackage{booktabs}

\usepackage[pagebackref=true,breaklinks=true,colorlinks,bookmarks=false]{hyperref}

\usepackage[capitalize]{cleveref}
\Crefname{section}{Sec.}{Secs.}
\crefname{section}{Sec.}{Secs.}
\Crefname{table}{Tab.}{Tabs.}
\crefname{table}{Tab.}{Tabs.}
\Crefname{figure}{Fig.}{Figs.}
\crefname{figure}{Fig.}{Figs.}
\Crefname{appendix}{App.}{Apps.}
\crefname{appendix}{App.}{Apps.}

\usepackage[utf8]{inputenc} 
\usepackage[T1]{fontenc}    
\usepackage{url}            
\usepackage{amsfonts}       
\usepackage{nicefrac}       
\usepackage{microtype}      
\usepackage{xcolor}         

\usepackage{epsfig}
\usepackage{tabularx}
\usepackage{bbm}
\usepackage{color}
\usepackage{wrapfig}
\usepackage{subcaption}
\usepackage{float}
\usepackage{makecell}
\usepackage[percent]{overpic}
\usepackage[numbers,compress]{natbib}
\usepackage{adjustbox}

\usepackage{multirow}
\usepackage{inconsolata}
\usepackage{comment}
\usepackage{xspace}

\newcommand{\stdhide}[1]{}
\newcommand{\std}[1]{\scriptsize{$\pm$#1}}

\newcommand{\ourdatareal}{OPDReal\xspace}
\newcommand{\ourdatacad}{OPDSynth\xspace}
\newcommand{\ourdatamulti}{OPDMulti\xspace}

\newcommand{\ourtask}{OPDMulti\xspace}

\newcommand{\opdformerbaseline}{\textsc{OpdFormer}\xspace}

\newcommand{\opdformerc}{\textsc{OpdFormer-C}\xspace}
\newcommand{\opdformero}{\textsc{OpdFormer-O}\xspace}
\newcommand{\opdformerp}{\textsc{OpdFormer-P}\xspace}

\newcommand{\opdrcnn}{\textsc{OPDRCNN}\xspace}
\newcommand{\opdrcnnc}{\textsc{OPDRCNN-C}\xspace}
\newcommand{\opdrcnno}{\textsc{OPDRCNN-O}\xspace}
\newcommand{\opdrcnnop}{\textsc{OPDRCNN-P}\xspace}
\newcommand{\opdnet}{\textsc{OpdRcnn}\xspace}

\newcommand{\drawer}{\texttt{drawer}\xspace}
\newcommand{\door}{\texttt{door}\xspace}
\newcommand{\lid}{\texttt{lid}\xspace}
\newcommand{\mttrans}{\texttt{prismatic}\xspace}
\newcommand{\mtrot}{\texttt{revolute}\xspace}

\newcommand{\axis}{\textbf{A}\xspace}
\newcommand{\orig}{\textbf{O}\xspace}
\newcommand{\mtype}{\textbf{M}\xspace}
\newcommand{\partdet}{\textbf{PDet}\xspace}

\newcommand{\mypara}[1]{\noindent\textbf{#1}}

\newcommand\best[1]{\textbf{#1}}

\newcolumntype{Y}{>{\centering\arraybackslash}X}
\newcommand\imgclip[2]{\adjincludegraphics[Clip={#1\width} {#1\height} {#1\width} {#1\height}]{#2}}

\begin{document}

\title{OPDMulti: Openable Part Detection for Multiple Objects}

\author{
Xiaohao Sun* \quad  Hanxiao Jiang* \quad Manolis Savva \quad Angel Xuan Chang \\
Simon Fraser University\\\href{https://3dlg-hcvc.github.io/OPDMulti/}{3dlg-hcvc.github.io/OPDMulti/}
}
\maketitle

\begin{abstract}
Openable part detection is the task of detecting the openable parts of an object in a single-view image, and predicting corresponding motion parameters. Prior work investigated the unrealistic setting where all input images only contain a single openable object. We generalize this task to scenes with multiple objects each potentially possessing openable parts, and create a corresponding dataset based on real-world scenes. We then address this more challenging scenario with OPDFormer: a part-aware transformer architecture. Our experiments show that the OPDFormer architecture significantly outperforms prior work. The more realistic multiple-object scenarios we investigated remain challenging for all methods, indicating opportunities for future work.
\end{abstract}
\section{Introduction}

Detecting the openable parts of real-world objects and predicting how the parts can move is useful in developing intelligent agents that can assist us with everyday household tasks.
Consider the simple task of `getting a spoon from the cabinet drawer'.
To achieve this, we need to identify what part of the cabinet is the drawer, that the drawer is openable, and that it opens with a translational motion.
Interest in tackling this problem has led to recent work focusing on mobility prediction of articulated object parts.

Prior work on mobility prediction aims to identify moving parts of an object, and predict the motion type and parameters of each moving part from a complete mesh~\cite{hu2018functionality} or 3D point cloud~\cite{wang2019shape2motion,hu2018functionality,yan2019rpm,shi2021self,shi2022p3}.
Recent work also considered mobility prediction from partial point clouds~\cite{li2020category} or depth images~\cite{jain2020screwnet,jain2022distributional}.
These methods rely mainly on depth information as input.
Another common limitation is strong category-specific assumptions or reliance on prior knowledge.
For instance, \citet{li2020category} assume a fixed kinematic chain (i.e. a separate model is trained for three-drawer cabinets vs two-drawer cabinets).

\begin{figure}[t]
\includegraphics[width=\linewidth]{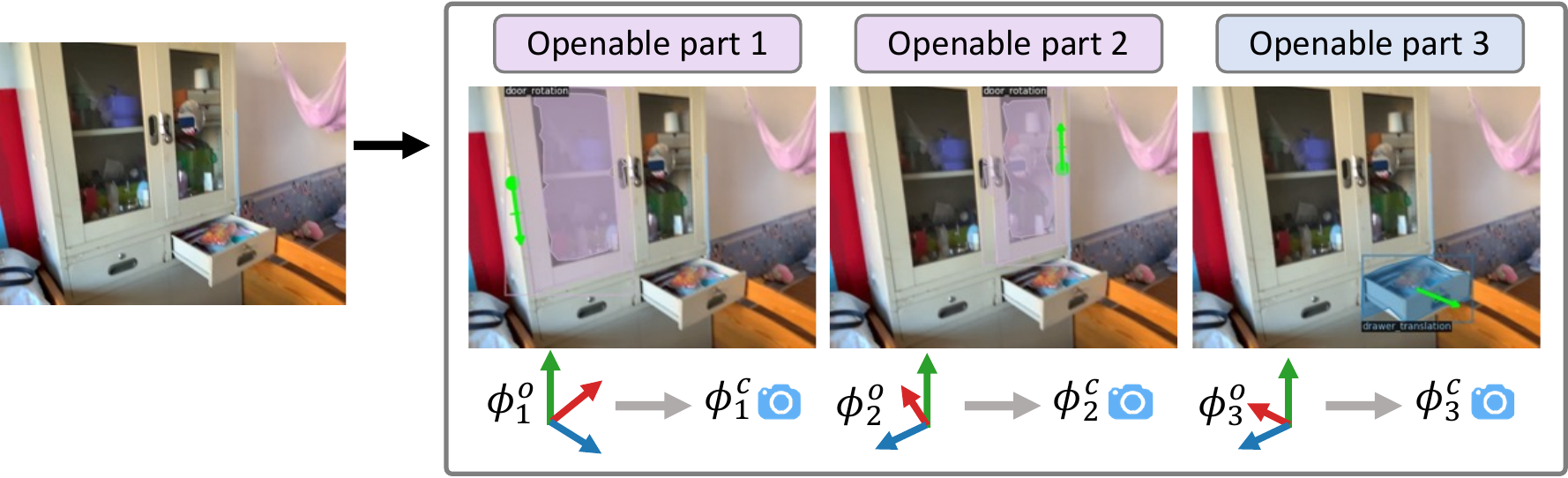}
\caption{We tackle the openable-part-detection (OPD) task: identifying parts that are openable and their motion parameters in a single-view image.
Our \opdformerbaseline architecture outputs segmentations for openable parts on potentially multiple objects, along with each part's motion parameters: motion type (translation or rotation, indicated by blue or purple mask), motion axis and origin (see green arrows and points).  For each openable part, we predict the motion parameters (axis and origin) in object coordinates ($\phi_i^o$) along with an object pose prediction to convert to camera coordinates ($\phi_i^c$).
}
\vspace{-15pt}
\label{fig:teaser}
\end{figure}

Recently, \citet{jiang2022opd} introduced the task of Openable Part Detection (OPD) where the openable parts and their corresponding motion parameters are predicted for a single articulated object from a single-view image (RGB, depth, or RGB-D).
This approach is object category agnostic, as it detects an arbitrary number of openable parts using Mask R-CNN~\cite{he2017mask} and predicts motion parameters for each part independently.
However, this work focus on single-object image and 
does not handle real-world scene layouts with potentially multiple objects, each with potentially multiple openable parts (e.g., real-world kitchens contain several cabinetry and drawer units).
To study OPD in real-world scenes with multiple objects, we introduce \ourtask, a challenging dataset of images with annotated part masks and motion parameters from real-world scenes containing multiple objects.
We create this dataset by leveraging recent work on articulated 3D scenes by \citet{mao2022multiscan}.

As noted in prior work~\cite{li2020category,jiang2022opd}, the motion parameters of openable parts (e.g., the direction in which a drawer slides open) is strongly correlated.
Since we have multiple objects in real scenes, we also need to model object pose for each part (in contrast to \citet{jiang2022opd} who only handle single objects).
We observe that parts in a object inform the movement of other parts in the same object, and that the pose of one object can inform the pose of another.
For instance, given the cabinets shown in \Cref{fig:teaser}, all the drawers move along the same axis, while the rotation axis of the doors is perpendicular to the translation axis of the drawers.  Objects are also likely to be placed either parallel or perpendicular to each other.
Thus, by leveraging features of other parts and the pose we can better detect and predict articulation parameters for each part.

We propose \opdformerbaseline, a part-informed transformer architecture leveraging self-attention to produce more globally consistent motion predictions.
The self-attention of this architecture better leverages the above observations of strong correlation between part positions, part mobility parameters, and object pose.
We compare three variants of our model: predicting directly in camera coordinates, predicting parts with a na\"ive single global pose, and with object pose predicted per-part.
We benchmark \opdformerbaseline against prior work and show that with the stronger architecture
and with per-part object pose prediction, we outperform prior methods by up to $10\%$ on openable part detection \& motion prediction with the same R50 backbone. Performance is further improved relative to baselines using a Swin-L backbone.

In summary, we make the following contributions: i) we construct a more realistic image-dataset for OPD with multiple objects ii) we propose a part-informed transformer architecture that leverages part--part and part--object pose correlations; iii) we systematically evaluate our approach and show it achieves state-of-the-art performance on the OPD task for both the single and multi-object setting.

\section{Related Work}

\mypara{2D instance segmentation.}
Instance segmentation in 2D is well-studied.
Before the popularity of vision transformers, prior work adopted region proposal-based methods~\cite{girshick2015fast,ren2015faster,he2017mask}.
\citet{carion2020end} used a transformer decoder to convert the instance segmentation task into a set prediction task with the Hungarian algorithm for a one-to-one matching loss.
MaskFormer~\cite{cheng2021per} further converted the problem into a mask classification problem to unify all 2D segmentation tasks (i.e. semantic segmentation, instance segmentation and panoptic segmentation) and achieved better results.
Recently, Mask2Former~\cite{cheng2021masked} achieved state-of-the-art results in 2D instance segmentation.
Our work builds on recent progress from instance segmentation, taking inspiration from transformer architectures that achieve state-of-the-art instance detection and segmentation performance.

\mypara{Articulated object understanding.}
With the increasing interest in embodied AI, understanding articulated objects is an important research direction.
A number of datasets of articulated objects have been recently introduced, including both synthetic \cite{xiang2020sapien,wang2019shape2motion} and scanned datasets \cite{jiang2022opd,qian2022understanding,liu2022akb,mao2022multiscan}.
These datasets have annotations of part segmentation and corresponding motion parameters.
Such data has enabled data-driven methods for predicting motion parameters from 3D meshes~\cite{hu2017learning} and points clouds~\cite{wang2019shape2motion,yan2019rpm}.
More recent work has focused on detecting articulated parts and their motion parameters from single-view point-clouds~\cite{li2020category}, images~\cite{zeng2021visual,jiang2022opd} and  videos~\cite{qian2022understanding,heppert2022category}, which are closer to real applications.
Researchers have also started to investigate how to use predicted segmentation and motion parameters to automatically create articulated objects~\cite{jiang2022ditto,collins20232}, including in scenes~\cite{hsu2023ditto}.

\mypara{Openable part detection.}
\citet{jiang2022opd} introduced the openable part detection (OPD) task to address the articulated object motion prediction problem for single-view image inputs.  In their work, they focused on images with a single main object and predicting the openable parts for that one object.
Our work generalizes the OPD task to more realistic images with multiple objects.
We also develop a part-informed transformer architecture that leverages object poses to predict more consistent and accurate part motions.

\section{\ourtask Task}

The OPD task seeks to identify all openable parts and their motion parameters from a single-view image $I$.
Specifically, to output a set of openable parts $P = \{p_1 \ldots p_k\}$ where an openable part is defined to be a drawer, door, or lid.
The output for each part is a segmentation mask $m_i$, 2D bounding box $b_i$, semantic label $l_i \in \{\drawer, \door, \lid\}$, and motion parameters $\phi_i$ specifying motion type $c_i \in \{\mttrans, \mtrot\}$, motion axis direction $a_i \in R^3$ and motion origin $o_i \in R^3$ (for revolute joints only).
\citet{jiang2022opd} focused on single-object images, which feature one object with at least one openable part.  
Here, we generalize the task and create a new dataset of real-world scenes with multiple objects, each possessing potentially multiple openable parts.
We call this new task and associated dataset \ourtask, reflecting the multi-object setting.

\subsection{\ourdatamulti dataset construction}
\label{sec:dataset}

\begin{figure*}[t]
\centering
\includegraphics[width=0.9\linewidth]{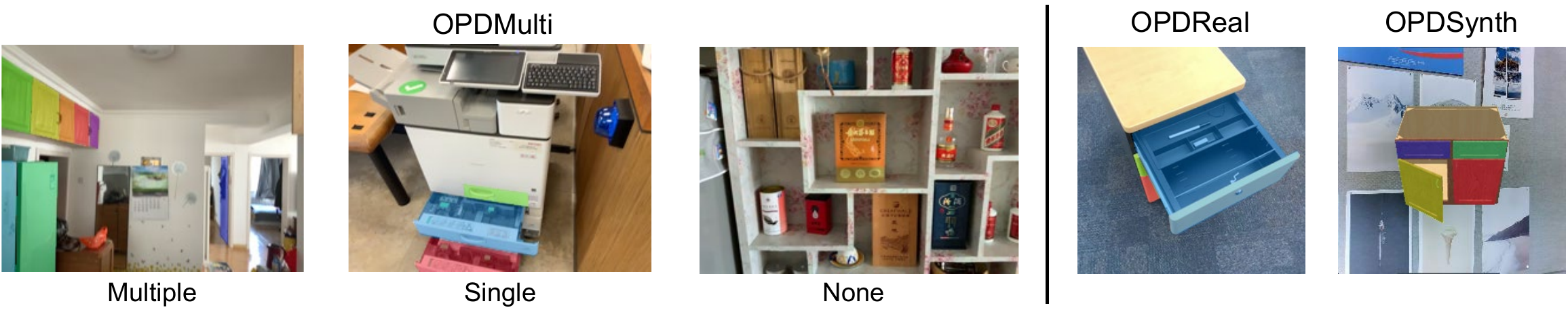}
\vspace{-8pt}
\caption{
Comparison of images from \ourdatamulti (left) to images from \ourdatareal~\cite{jiang2022opd} and \ourdatacad~\cite{jiang2022opd}.
Different mask colors indicate different openable parts.
Our \ourdatamulti dataset is more realistic and diverse with images from varied viewpoints and with multiple/single/no openable objects.
}
\label{fig:data-compare}
\end{figure*}

\begin{table}
\centering
\resizebox{\linewidth}{!}
{
\begin{tabular}{@{} ccc rrrr @{}}
\toprule
dataset & type & obj per frame & objects & categories & parts  & frames
\\
\midrule
\ourdatacad & synth & 1 & 683 & 11 & 1343 & 100K  \\
\ourdatareal & real & 1 & 284 & 8 & 875 & 30K \\
\ourdatamulti & real & 0/1/1+ & 217 (4973) & 33 (458) & 688 (4387) & 64K \\
\bottomrule
\end{tabular}
}
\caption{
Statistics comparing our \ourdatamulti dataset with \ourdatacad and \ourdatareal from \citet{jiang2022opd}.  Since \ourdatamulti contains objects in scenes, we report both the number of articulated objects and total objects (in parentheses).
}
\label{tab:data-opdmulti-stats-compare}
\end{table}
\begin{figure}[t]
\centering
\begin{overpic}[width=\linewidth]{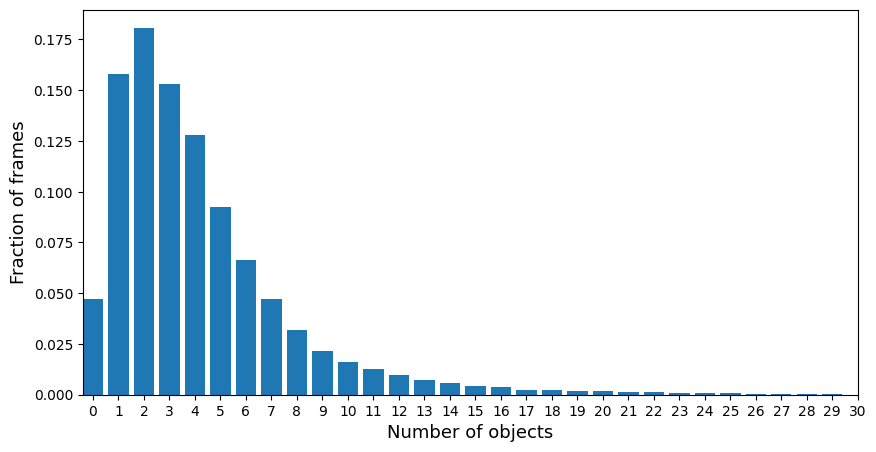}
\put(32,15){\includegraphics[width=.65\linewidth]{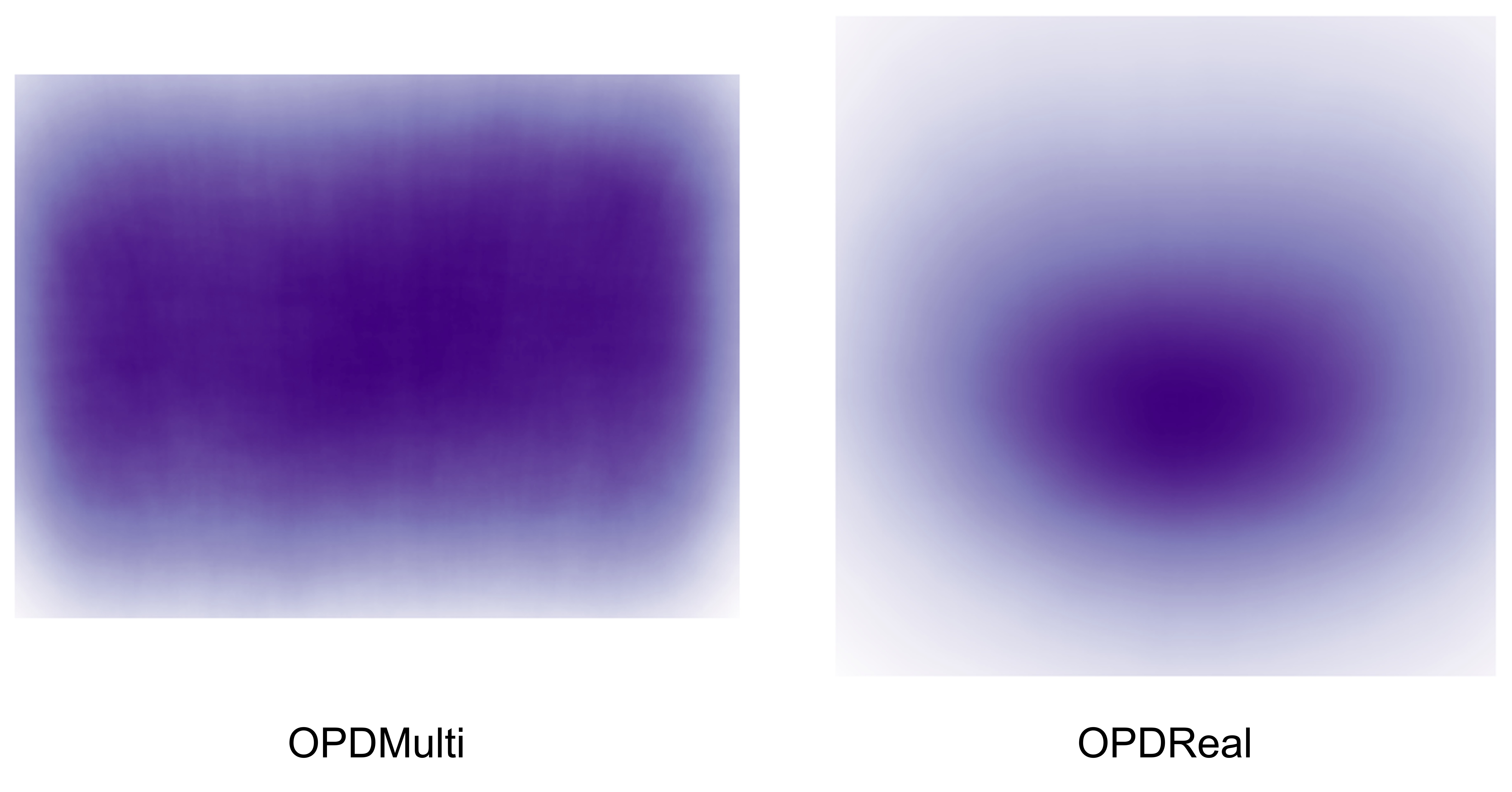}}
\end{overpic}
\vspace{-20pt}
\caption{
Distribution of frames over number of objects observed for OPDMulti.
Inset: distribution of openable part pixels aggregated across frames.
We see that the distribution of openable part mask pixels in OPDMulti is more uniformly spread out compared to OPDReal, where the openable part is mostly in the center.
Note that OPDReal has center-cropped images whereas in OPDMulti we preserve the original image resolution and aspect ratio.
}
\label{fig:frame-distribution}
\end{figure}
\begin{table}
\centering
{
\begin{tabular}{@{} r rrrr r @{}}
\toprule
split & frames & none & single & multiple & parts/frame
\\
\midrule
train & 44002 & 31189 & 10911 & 1902 & 1.64 \\
val & 10168 & 6424 & 3055 & 689 & 1.85 \\
test & 10043 & 6818 & 2732 & 493 & 1.80 \\
\midrule
total & 64213 & 44431 & 16698 & 3084 & 1.71 \\
\bottomrule
\end{tabular}
}
\vspace{-5pt}
\caption{
Number of frames in \ourtask with no/single/multiple openable objects.  Parts/frame is the average for frames with at least one openable part.
}
\label{tab:data-opdmulti-stats}
\end{table}

To create \ourtask image dataset, we leverage MultiScan~\cite{mao2022multiscan}, a dataset of RGB-D reconstructions of real indoor scenes providing object and part-level annotations.
We use the RGB-D video frames in this dataset along with part and part articulation annotations to create our \ourdatamulti dataset.

Specifically, we sample frames from RGB videos in the MultiScan dataset and project object and part segmentation masks to the image plane.
We also process the annotated motion parameters and object poses to the same format as the \ourdatacad and \ourdatareal datasets~\cite{jiang2022opd}.
Unlike prior work, we keep the full image resolution instead of center-cropping to avoid dropping objects that appear on the sides.
Since some frames may contain small or partial parts that are cropped and hard to detect, we ignore openable part annotations that cover less than $5\%$ of the image pixels.
Overall, we use 273 scans from 116 MultiScan scenes to create our image dataset, following the MultiScan train/val/test set split.  

We find that some of the projected annotations are noisy and inaccurate.
To ensure that our evaluation dataset is of high quality, we manually inspect all frames in the val and test splits and indicate whether they have \emph{mask} or \emph{motion} errors.
Mask errors are typically caused by reconstruction issues (e.g., a door with glass panes is not fully reconstructed so when projected onto the image the annotated mask is incomplete).
We also observe shifts in the mask for some frames if the estimated camera poses are not consistent with the final reconstruction. 
We find that 512 val set and 1749 test set openable part mask annotations are noisy (out of a total of 6077 val and 4704 test openable parts).
For mask error cases, we manually correct the mask using the Toronto Annotation Suite~\cite{torontoannotsuite}.
See the supplement for details.

\subsection{\ourdatamulti dataset statistics}

Following \ourdatacad and \ourdatareal~\cite{jiang2022opd}, we focus on three openable part types (drawer, door, lid) that are common across many object categories.
\Cref{tab:data-opdmulti-stats-compare,tab:data-opdmulti-stats} provides dataset statistics.
Our \ourdatamulti contains 33 object categories with at least a door, drawer, or lid (23, 15, 8 categories respectively).
Example categories include cabinets, refrigerators, wardrobes, microwaves, washing machines, nightstands, toilets, printers, and rice cookers, with a long-tailed distribution from frequent (182 cabinets) to infrequent (7 rice cookers).
Since our focus is on rigid openable objects, we do not include non-rigid object such as bags in our dataset.

In \Cref{fig:data-compare} we compare the images in \ourdatamulti vs prior datasets.
Note that images from \ourdatamulti are more varied with frames showing a variable number of openable objects including multiple openable objects, single openable object with natural background clutter, and frames with no openable objects.
\Cref{fig:frame-distribution} shows the distribution over number of objects and location of part pixels for the images in the resulting dataset.
From the inset, which shows the distribution of openable part pixels aggregated across the frames, we can see that \ourdatamulti has a broader part pixel distribution than \ourdatareal.
For \ourdatareal, most of the openable parts are in the center while in \ourdatamulti the openable parts are spread more evenly across the frame.
Overall, the images in \ourdatamulti are more diverse with both distant and close-up views and views from different angles.
\begin{figure*}[t]
\includegraphics[width=\linewidth]{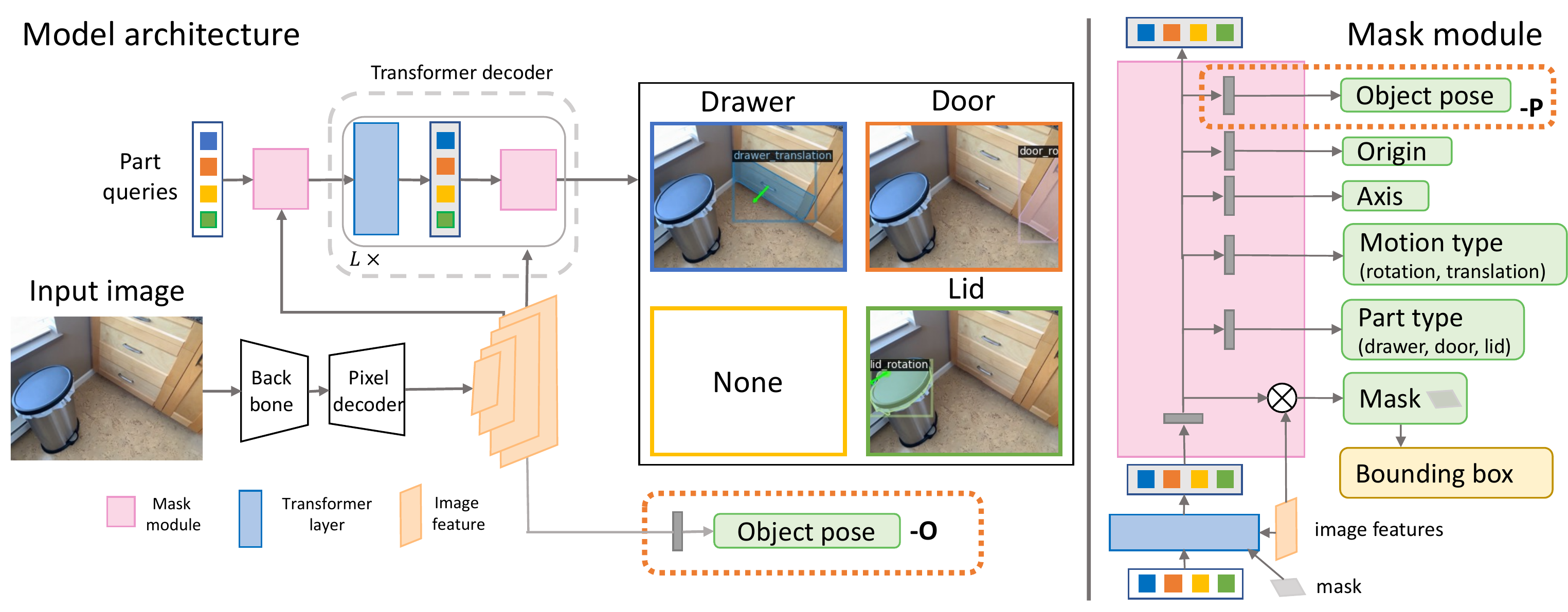}
\vspace{-15pt}
\caption{
Our \opdformerbaseline architecture is based on the Mask2Former~\cite{cheng2021masked} architecture.
The left side shows the overall network while the right shows the mask module in detail.  The Mask2Former employs an image backbone and pixel decoder to obtain pixel-level embeddings, which are passed to a transformer decoder with masked attention together with learnable part queries to learn embeddings that are used to predict the part type and mask (by the mask module).
To obtain a high-resolution mask, Mask2Former uses a multi-scale strategy with visual feature maps at increasing resolutions, each of which are fed into the transformer decoder.  The transformer decoder unit is then stacked for $L$ layers.
We enhance the mask module to predict the part motion parameters (motion type, origin, axis) in addition to the part type and mask (see green boxes). The part bounding box is computed directly from the mask.
We investigate three variants of the architecture that predict motion parameters either directly in camera coordinates (-C), or in object coordinates which are then transformed to camera coordinates via a global pose or a per-part object pose.
The pose prediction variants are indicated in \textcolor{orange}{orange} dashed boxes: global pose (`-O' at middle bottom), and per-part object pose (`-P' top right).
The detected parts in the center correspond to the part queries.
}
\label{fig:network-structure}
\end{figure*}

\section{Approach}

We adopt the detect-and-predict strategy for openable part detection and motion parameter estimation, following \citet{jiang2022opd}.
Our architecture replaces the Mask R-CNN~\cite{he2017mask} detection component with Mask2Former~\cite{cheng2021masked}. 
Mask2Former uses the transformer decoder to predict instance masks and classes, matching against ground truth using the Hungarian algorithm during training.
We extend the Mask2Former architecture to predict part motion parameters in the mask module, and create three variants of the architecture that predict the motion parameters using different coordinate frames.
Our key differences from \citet{jiang2022opd} are that: 1) we replace the MaskRCNN segmentation architecture with Mask2Former; and 2) we introduce a per-part object pose prediction (instead of a global object pose prediction).

\subsection{Model variants}

All models eventually predict motion parameters in camera coordinates (\textbf{C}).
However, as noted in prior work~\cite{li2020category,jiang2022opd}, it is useful to predict motion parameters in the object coordinate frame as the motion axes are often parallel to one of the main axes of the object (see supplement).
The object pose is used as a bridge to transform between the object coordinate frame and the camera coordinate frame.

\citet{jiang2022opd} used the entire image to predict a single object pose, ignoring the fact that there could be multiple objects with different poses. 
To alleviate this limitation, we develop two variants of our architecture for pose prediction, predicting a single \emph{global} pose vs predicting a different object pose for each \emph{part}.
By predicting the object pose per part, we can handle multiple objects without explicitly detecting each object.
This allows us to have an object-agnostic method that can generalize across object categories.
In addition, we consider a base variant that predicts directly in the camera coordinates.

\mypara{Camera coordinates.} 
The base variant \opdformerc does not predict the object pose, and predict the motion parameters in the camera coordinate directly (see \Cref{fig:network-structure}, without orange dashed boxes).  Note that it is the direct analogue of the Mask R-CNN based baseline \opdrcnn-C from \citet{jiang2022opd}.

\mypara{Single global pose.}
In this variant, we predict a single global pose for all objects and parts in the input image.
This is predicted directly from the image features of the entire image (see \Cref{fig:network-structure} dashed box with label `-O').
We call this variant \opdformero as it is the direct analogue of the OPDRCNN-O introduced in OPD~\cite{jiang2022opd}.
For \ourdatamulti, we train with the scene coordinates defining a global pose, and transform relevant motion parameters from camera coordinates to these scene coordinates.

\mypara{Per-part object pose.}
When there are multiple objects in an image, each of the objects can have a different pose and its openable parts would have motion parameters strongly correlated with that object's pose.
To account for this, we add an additional head for each part that predicts the object's pose (see \Cref{fig:network-structure} dashed box with `-P' label).
We call this variant \opdformerp and compare it against \opdrcnnop, an extension of the MaskRCNN based model from \citet{jiang2022opd} to predict per-part object pose.
For \ourdatamulti, we leverage the object oriented bounding boxes in MultiScan~\cite{mao2022multiscan} to obtain object poses and transform motion parameters to object coordinates for training.

\mypara{Parameterization.}
In all variants, the motion parameters and object pose are parameterized in the same way as \citet{jiang2022opd}'s \opdrcnn: motion axis and motion origin are 3-dim vectors and object pose is a 12-dim vector (9 for rotation and 3 for translation).
Motion type prediction is trained with a cross-entropy loss and other motion parameters and object pose use a smooth L1 loss with $\beta=1$.

\subsection{Network Architecture and Losses}

The overall architecture and mask module with per-part prediction heads are shown in the left and right sides of \Cref{fig:network-structure}.
Our architecture uses a Mask2Former module for part segmentation and self-attention over parts.
We use the same multiscale pixel decoder and transformer decoder (with 100 queries) as in \citet{cheng2021masked}, and an R50 backbone for fair comparison with \citet{jiang2022opd}'s \opdnet.

For the segmentation and motion losses, we add the auxiliary loss after each transformer decoder.
The object pose loss is determined by the specific architecture variant and is either a single loss term or one loss term per part.

\mypara{Segmentation losses.}
We use the same set of losses as Mask2Former~\cite{cheng2021masked}, including the binary cross-entropy loss ($L_\text{ce}$) and the dice loss ($L_\text{dice}$)~\cite{milletari2016v} for the mask segmentation, and cross-entropy loss ($L_\text{cls}$) for the mask classification: $L_\text{seg} = \lambda_{ce} L_\text{ce} + \lambda_{dice} L_\text{dice} + \lambda_{cls} L_\text{cls}$.
We adopt the loss weights proposed in Mask2Former, $\lambda_{ce} = 5, \lambda_{dice} = 5$ and $\lambda_{cls} = 2$ for matched predictions and $0.1$ for unmatched.

\mypara{Motion losses.}
Motion prediction losses are based on \opdrcnn~\cite{jiang2022opd}.
We use a cross entropy loss for the motion type ($L_\text{c}$), combined with smooth L1 losses for the motion axis ($L_\text{a}$) and motion origin ($L_\text{o}$): $L_\text{mot} = \lambda_c L_c + \lambda_a L_a + \lambda_o L_o$.
We also use the same loss weight ratios.
Specifically, we set $\lambda_c = 2, \lambda_a = 16, \lambda_o = 16$ for our experiments.

\mypara{Object pose loss.}
Object pose prediction is trained under the smooth L1 loss ($L_\text{pose}$) with $\lambda_\text{pose} = 30$.

We sum all of the above losses to obtain the overall loss used during training: $L = L_\text{seg} + L_\text{mot} + \lambda_\text{pose} L_\text{pose}$.

\section{Experiments}

We compare our proposed architecture against baselines on both single object datasets (\ourdatacad, \ourdatareal from \citet{jiang2022opd}) and the new multiple object dataset we created (\ourdatamulti).
We also conduct an analysis of part consistency on single objects and the challenges of handling multiple objects.
In the main paper, we present experiments for RGB input images.
See the supplement for results with depth only (D) and RGBD, and additional analysis.

\subsection{Implementation details} 
Our architecture is based on Mask2Former~\cite{cheng2021masked} as implemented in Detectron2~\cite{wu2019detectron2}.
We use the R-50 backbone Mask2Former model pretrained on COCO~\cite{lin2014microsoft} instance segmentation to initialize our weights, and train with AdamW~\cite{loshchilov2017decoupled}.
The learning rate and other hyperparameters match those used by Mask2Former.
Our experiments are carried out on a machine with 64GB RAM and an RTX 2080Ti GPU.
We train each model end-to-end for $60000$ steps and pick the best checkpoint based on val set performance (+\mtype{}\axis{}\orig).
Models evaluated on \ourtask are first pretrained on \ourdatareal and then finetuned on the \ourdatamulti train split.
The \opdrcnn baselines are first pretrained on \ourdatacad, then \ourdatareal, and finally \ourdatamulti.

We note that the predicted object pose rotation matrix is not guaranteed to be a valid rotation matrix. \citet{jiang2022opd} did not address this issue.
We convert the predicted rotation matrix into a unit quaternion and back using PyTorch3D~\cite{ravi2020accelerating} to ensure a valid rotation.
The results for \opdnet are approximately the same as without such post-processing.
For \ourdatamulti, we use a confidence threshold of $0.8$ to determine whether a predicted part is valid.

\subsection{Experimental setup}

For single objects, we evaluate our method on two datasets introduced in OPD \cite{jiang2022opd}, \ourdatacad and \ourdatareal.
For multiple objects, we evaluate on \ourdatamulti.

\mypara{Metrics.}
We use the evaluation metrics for part detection and motion prediction from \citet{jiang2022opd}. 
The metrics extend the traditional mAP metric for detection to the motion prediction task, including two main metrics: mAP@IoU=0.5 for the predicted part label and 2D bounding box (\partdet).
For each metric, the detection is further constrained by whether: motion type is matched (+\mtype{}), motion type and motion axis are matched (+\mtype{}\axis{}), and whether motion type, axis and origin are all matched (+\mtype{}\axis{}\orig), within predefined error thresholds.
Note that \citet{jiang2022opd}'s metrics were only defined for inputs with openable parts.
Since we have frames with no openable parts, we measure the percentage of those we correctly predicted as having no openable parts.

\mypara{Methods.}
We compare variants of our \opdformerbaseline with the MaskRCNN-based \opdrcnn~\cite{jiang2022opd}.
We compare the following variants: predicting directly in camera coordinates (\textsc{-C}), vs predicting a single global pose (\textsc{-O}) vs predicting per-part object poses (\textsc{-P}). 

\begin{table}
\resizebox{\linewidth}{!}
{
\begin{tabular}{@{} ll rrrr @{}}
\toprule
& & \multicolumn{4}{c}{Part-averaged mAP $\% \uparrow$} \\
\cmidrule(l{0pt}r{2pt}){3-6}
Dataset & Model & \partdet & +\mtype & +\mtype{}\axis & +\mtype{}\axis{}\orig \\
\midrule
\multirow{7}{*}{\ourdatacad}
& \opdrcnnc~\cite{jiang2022opd} & 74.3\std{0.27} & 72.3\std{0.29} & 40.2\std{0.09} & 36.5\std{0.17} \\
& \opdrcnno~\cite{jiang2022opd} & 74.2\std{0.34} & 72.4\std{0.32} & 52.4\std{0.27} & 47.0\std{0.36} \\
& \opdrcnnop & 73.2\std{0.64} & 71.2\std{0.69} & 51.6\std{0.47} & 44.8\std{0.32} \\
& \opdformerc & 77.3\std{0.40} & 74.9\std{0.42} & 48.9\std{0.23} & 43.9\std{0.09} \\
& \opdformero & 77.8\std{0.54} & 75.7\std{0.47} & 57.5\std{0.15} & 52.4\std{0.35} \\
& \opdformerp & \best{79.0}\std{0.23} & \best{76.7}\std{0.23} & \best{58.6}\std{0.94} & \best{53.4}\std{0.28} \\
\midrule
\multirow{7}{*}{\ourdatareal}
& \opdrcnnc~\cite{jiang2022opd} & 57.6\std{0.10} & 55.5\std{0.24} & 15.6\std{0.28} & 14.7\std{0.29}\\
& \opdrcnno~\cite{jiang2022opd} & 57.0\std{0.49} & 54.7\std{0.57} & 27.9\std{0.49} & 25.7\std{0.41} \\
& \opdrcnnop & 57.6\std{0.62} & 54.7\std{0.59} & 26.9\std{0.03} & 25.1\std{0.19} \\
& \opdformerc & 57.9\std{1.31} & 56.0\std{1.09} & 29.7\std{0.51} & 28.3\std{0.49} \\
& \opdformero & \best{61.8}\std{0.58} & \best{59.4}\std{0.55} & 31.2\std{0.58} & 28.9\std{0.57} \\
& \opdformerp  & 58.8\std{0.66} & 56.2\std{0.58} & \best{35.4}\std{0.20} & \best{33.7}\std{0.18} \\
\midrule
\multirow{7}{*}{\ourdatamulti}
& \opdrcnnc~\cite{jiang2022opd} & 27.3\std{0.10} & 25.7\std{0.10} & 8.8\std{0.25} & 7.8\std{0.20} \\
& \opdrcnno~\cite{jiang2022opd} & 20.2\std{0.42} & 18.3\std{0.62} & 3.9\std{0.07} & 0.5\std{0.12} \\
& \opdrcnnop & 20.9\std{0.44} & 19\std{0.35} & 7.2\std{0.25} & 5.7\std{0.22} \\
& \opdformerc & 30.3\std{1.02} & 28.9\std{0.99} & 13.1\std{0.55} & 12.1\std{0.49} \\
& \opdformero & 30.1\std{0.15} & 28.5\std{0.18} & 5.2\std{0.10} & 1.6\std{0.03} \\
& \opdformerp & \best{32.9}\std{0.69} & \best{31.6}\std{0.72} & \best{19.4}\std{0.38} & \best{16.0}\std{0.03} \\
\bottomrule
\end{tabular}
}
\caption{Comparison of \opdrcnn and \opdformerbaseline on validation set RGB input images for the three datasets (\ourdatacad and \ourdatareal for single-object, and \ourdatamulti for multiple-object real scenes).
Our \opdformerbaseline variants outperform baselines especially on the multi-object inputs from \ourdatamulti.}
\label{tab:results-OPDAll-val-mini}
\end{table}

\begin{figure*}
\centering
\setkeys{Gin}{width=\linewidth}
\begin{tabularx}{\linewidth}{Y Y Y Y Y Y}
\toprule
\small{GT} &
\imgclip{0}{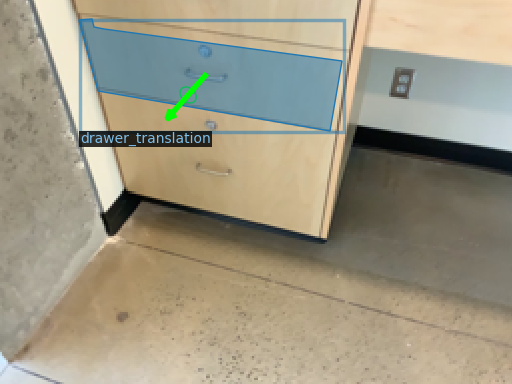} & 
\imgclip{0}{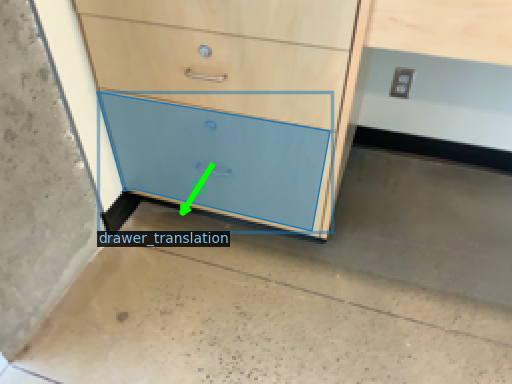} &
\imgclip{0}{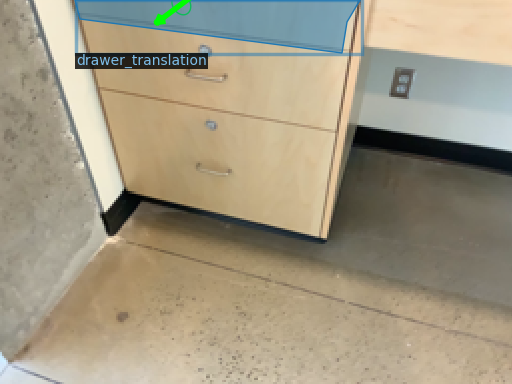} &
\imgclip{0}{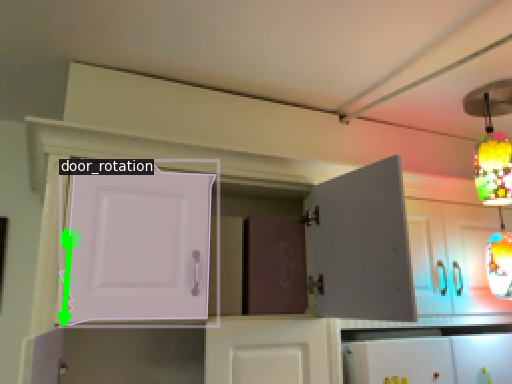} &
\imgclip{0}{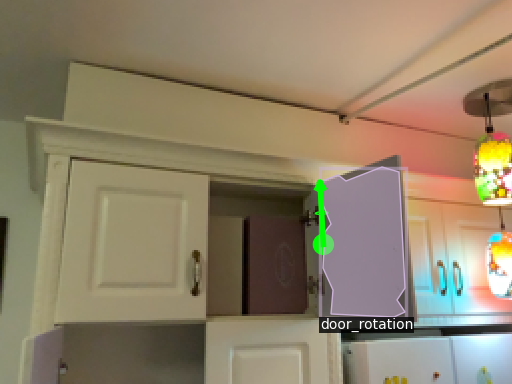}\\

\small{\opdrcnnop} &
\imgclip{0}{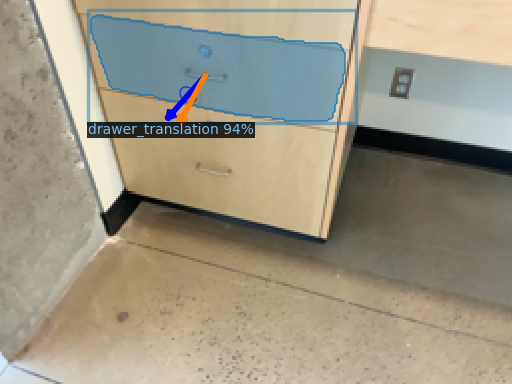} & 
\imgclip{0}{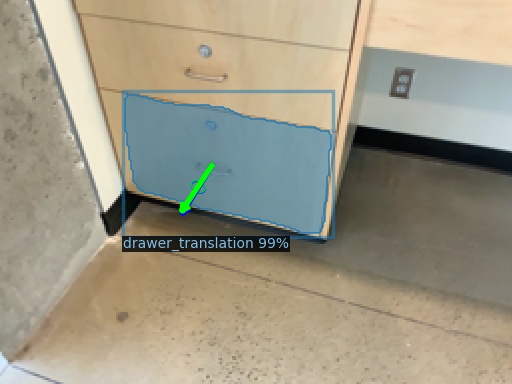} &
Miss &
\imgclip{0}{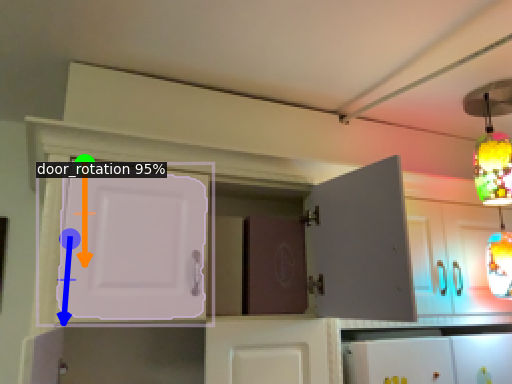} &
\imgclip{0}{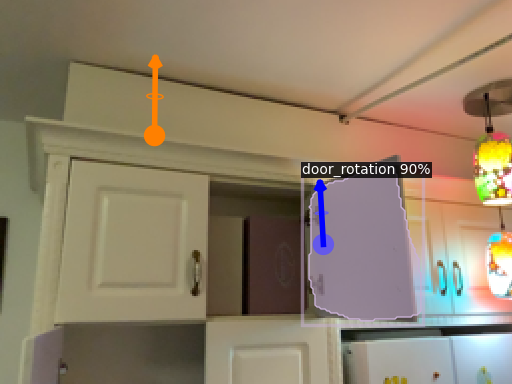}\\
\small{Axis (origin) error} & 7.537 & 2.098 & - & 6.078 (0.067) & 8.752 (0.149)\\

\small{\opdformerp} &
\imgclip{0}{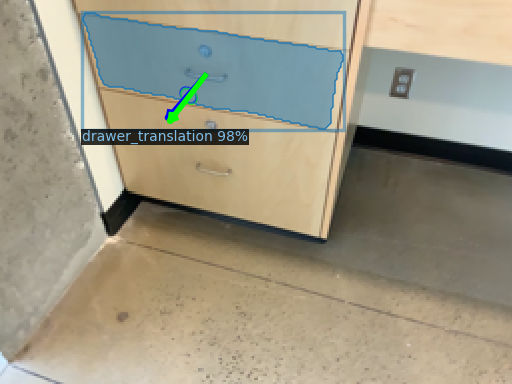} & 
\imgclip{0}{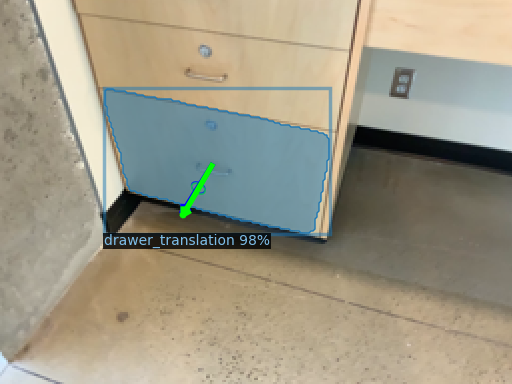} &
\imgclip{0}{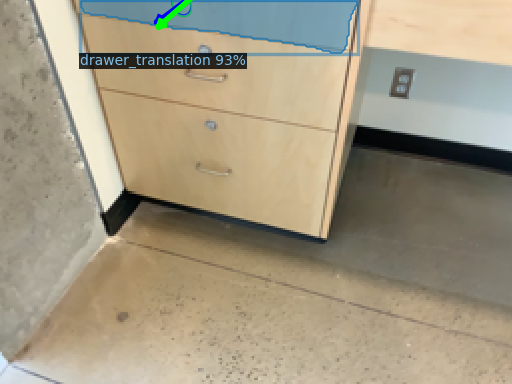} &
\imgclip{0}{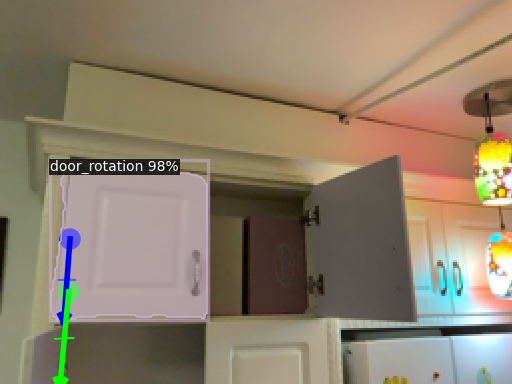} &
\imgclip{0}{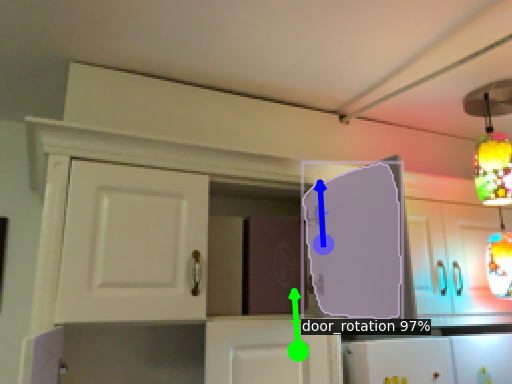}\\
\small{Axis (origin) error} & 2.131 & 2.956 & 2.153 & 3.247 (0.019) & 1.975 (0.06)\\

\midrule

\small{GT} &
\imgclip{0}{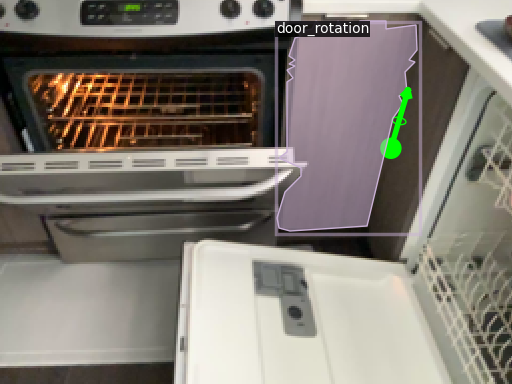} & 
\imgclip{0}{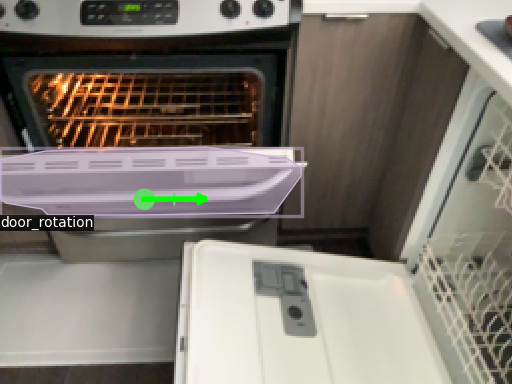} &
\imgclip{0}{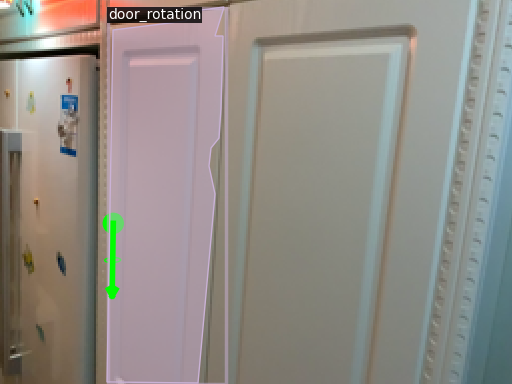} &
\imgclip{0}{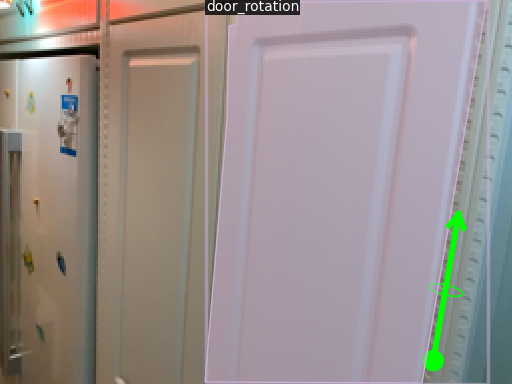} &
\imgclip{0}{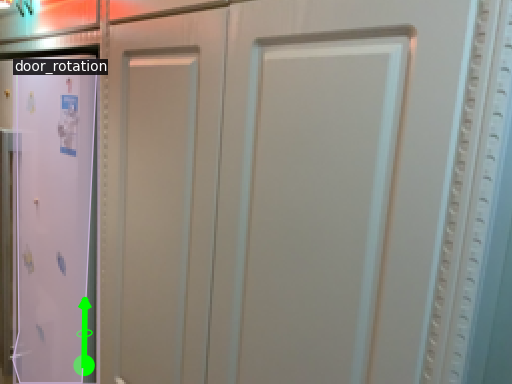}\\

\small{\opdrcnnop} &
\imgclip{0}{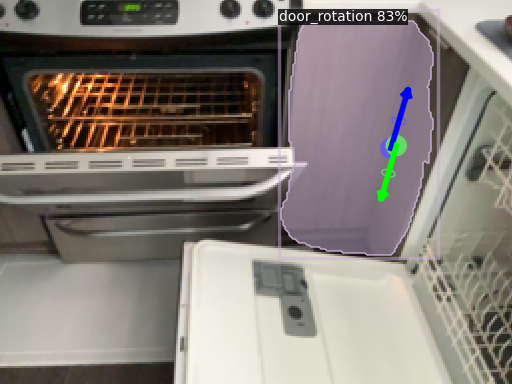} & 
Miss &
\imgclip{0}{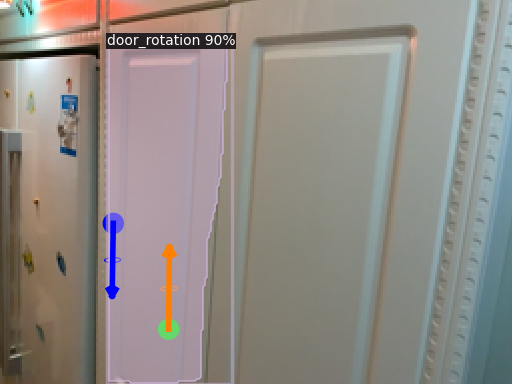} &
\imgclip{0}{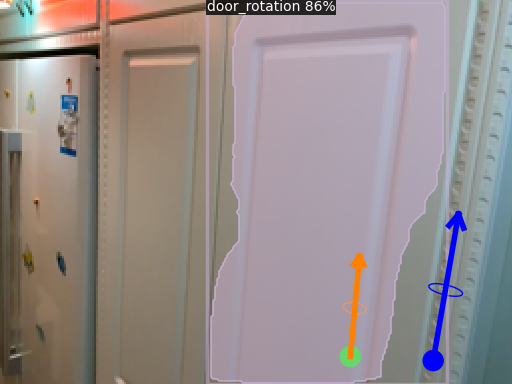} &
\imgclip{0}{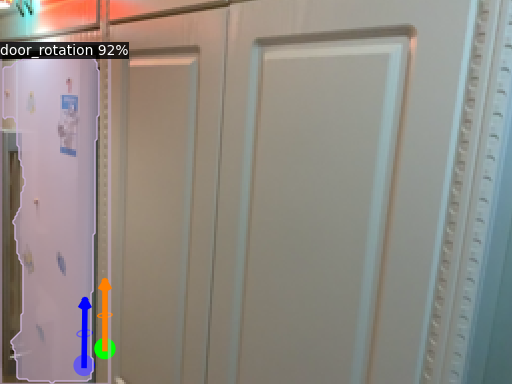}\\
\small{Axis (origin) error} & 4.458 (0.013) & - & 6.329 (0.057) & 5.757 (0.056) & 8.305 (0.029)\\

\small{\opdformerp} &
\imgclip{0}{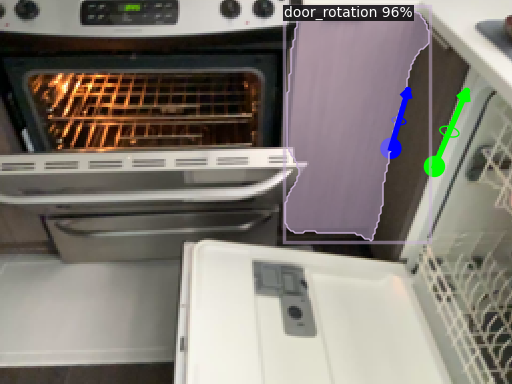} & 
\imgclip{0}{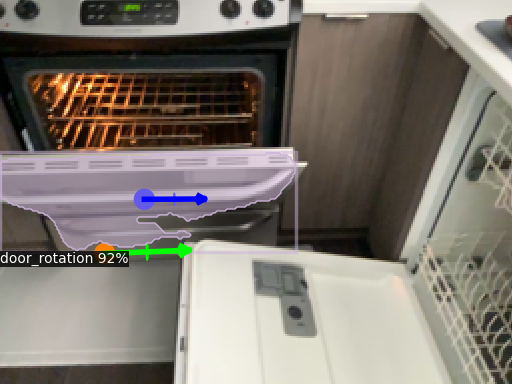} &
\imgclip{0}{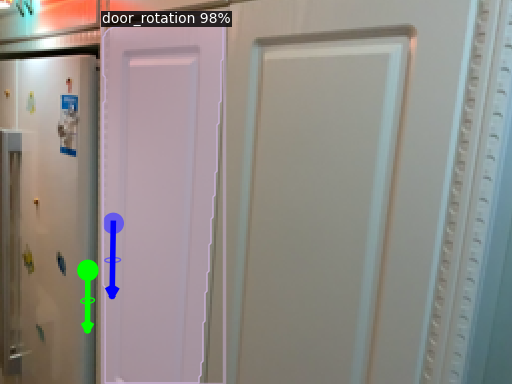} &
\imgclip{0}{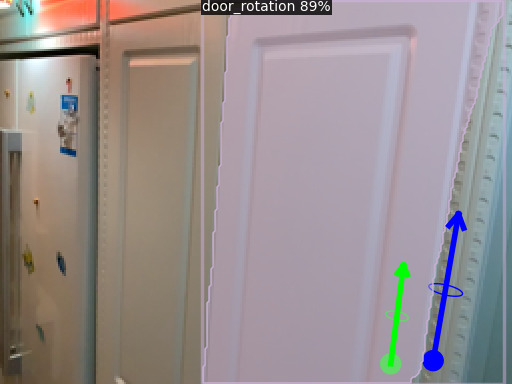} &
\imgclip{0}{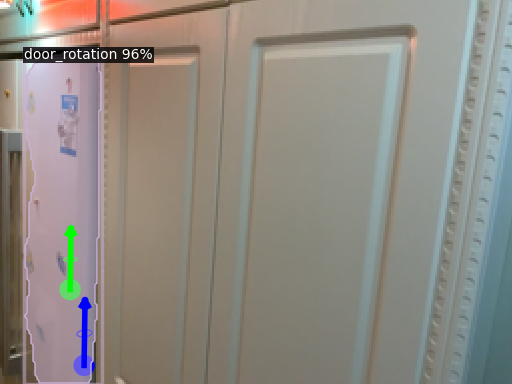}\\
\small{Axis (origin) error} & 2.021 (0.087) & 4.099(0.234) & 1.038 (0.058) & 1.619 (0.063) & 1.569 (0.096)\\

\bottomrule
\end{tabularx}
\caption{
Example predictions on the \ourdatamulti val split. 
The first row in each group is the ground truth (GT) with the motion axis in green.  
The following rows are predictions from \opdrcnnop and \opdformerp with axis error and origin error indicated if the motion type is rotation.
The GT axis is in \textcolor{blue}{blue} and the predicted axis is in \textcolor{green}{green} if it is within $5^\circ$ of the GT, \textcolor{orange}{orange} if between $5^\circ$ and $10^\circ$, and \textcolor{red}{red} if the angle difference is greater than $10^\circ$.
The axis origin is visualized with the same color scheme using error thresholds of 0.1 and 0.25.
Overall, \opdformerp provides significantly more accurate openable part predictions, in particular for the scenarios in the bottom that contain multiple objects or multiple parts.
}
\label{fig:vis-compare-real-multi}
\end{figure*}

\subsection{Results}

\Cref{tab:results-OPDAll-val-mini} evaluates the different methods on RGB input images from the val set of the three datasets.  We report the mean and standard error across three runs with different seeds. 
See the supplement for depth and RGB-D input results, motion averaged metrics, and for performance on the test set.
\Cref{fig:vis-compare-real-multi} shows example predictions on \ourdatamulti, and the supplement provides qualitative results on \ourdatacad and \ourdatareal.

Our \opdformerbaseline variants outperform the \opdrcnn baselines on all metrics.
One reason is the stronger part detection (\partdet) provided by the Mask2Former backbone.
We note that the \opdformerbaseline variants with the R50 backbone actually have fewer parameters than \opdrcnn methods, indicating that the performance gains are not due to increased parameters.
For example, \opdrcnnop has 46.1M parameters whereas \opdformerp has 42.0M parameters.
This observation is similar for other \opdformerbaseline variants and corresponding \opdrcnn baselines (see supplement).

\mypara{Are camera coordinates useful?}
As observed in \citet{jiang2022opd}, predicting motion parameters in object coordinates and predicting the object pose (\opdrcnno) outperform prediction in camera coordinates (\opdrcnnc).
This is true for the single object case (\ourdatacad and \ourdatareal), but not for \ourdatamulti where the assumption of one global object coordinate does not hold.

\mypara{Is having per-part object pose prediction important?}
When we take motion parameters into account, we see the advantage of per-part object pose predictions with the transformer-based architecture (\opdformerp).
On the main metric (+\mtype{}\axis{}\orig), our per-part \opdformerp consistently outperforms the global \opdformero, which in turn outperforms \opdrcnno by \citet{jiang2022opd}.
Interestingly, \opdrcnnop does not help over \opdrcnno for the single object scenario.

\begin{table}
\resizebox{\linewidth}{!}{
\begin{tabular}{@{} l r rr rr @{}}
\toprule
 & No AO $\% \uparrow$ & \multicolumn{2}{c}{Single AO $\% \uparrow$} & \multicolumn{2}{c}{Multiple AO $\% \uparrow$} \\
\cmidrule(l{0pt}r{2pt}){2-2} \cmidrule(l{2pt}r{0pt}){3-4} \cmidrule(l{2pt}r{0pt}){5-6}
Model & \textbf{Accuracy} & \partdet & +\mtype{}\axis{}\orig & \partdet & +\mtype{}\axis{}\orig  \\
\midrule
\opdrcnnc~\cite{jiang2022opd}  & \best{58.6} & 43.6 & 11.8 & 37.6 & 8.6\\
\opdrcnno~\cite{jiang2022opd}  & 57.5 & 34.8 & 0.5 & 30.5 & 0.4\\
\opdrcnnop & 50.8 & 40.0 & 10.0 & 34.0 & 8.9 \\
\opdformerc & 27.3 & 60.1 & 21.9 & 36.1 & 14.6 \\
\opdformero  & 16.7 & 59.4 & 2.5 & 35.8 & 1.5\\
\opdformerp & 35.0 & \best{61.4} & \best{28.7} & \best{40.2} & \best{15.2} \\
\bottomrule
\end{tabular}
}
\caption{We compare the performance of the models for images with no/one/multiple articulated objects (AO) on the \ourdatamulti validation set.  For `No AO', we compute the percent of frames for which the method correctly predicted there was no openable parts.
}
\label{tab:results-OPDMulti-split}
\end{table}

\begin{table}[t]
\resizebox{\linewidth}{!}{
\begin{tabular}{@{} ll rr rr @{}}
\toprule
& & \multicolumn{2}{c}{Pose Rotation} & \multicolumn{2}{c}{Pose Translation} \\
\cmidrule(l{0pt}r{2pt}){3-4} \cmidrule(l{2pt}r{0pt}){5-6}
Dataset & Model & MedErr $\downarrow$ & Acc:5 $\uparrow$ & MedErr $\downarrow$ & Acc:0.1 $\uparrow$ \\
\midrule
\multirow{2}{*}{\ourdatacad} 
& \opdrcnnop & 4.28 & 0.58 & 0.16 & 0.28 \\
& \opdformerp &  \best{2.47} & \best{0.78} & \best{0.11} & \best{0.46}  \\
\midrule
\multirow{2}{*}{\ourdatareal} 
& \opdrcnnop & 8.33 & 0.23 & 0.19 & 0.16 \\
& \opdformerp & \best{4.96} & \best{0.51} & \best{0.14} & \best{0.29} \\
\midrule
\multirow{2}{*}{\ourdatamulti} 
& \opdrcnnop & 19.86 & 0.05 & 0.27 & 0.06 \\
& \opdformerp & \best{8.09} & \best{0.27} & \best{0.21} & \best{0.12} \\
\bottomrule
\end{tabular}
}
\caption{Object pose error on the val set for all three datasets. Rotation error is in degrees and translation error is normalized by the diagonal length of the object. For accuracy, we use thresholds of $5^\circ$ for rotation and 0.1 (of object diagonal) for translation. Averages are computed part-wise. Accuracy counts matched pairs of GT and prediction in the same way as the mAP@50 metric, with higher confidence masks picking GT first and IoU = 50 threshold.}
\label{tab:results-pose-error-mini}
\end{table}

\mypara{How challenging is \ourdatamulti?}
\ourdatamulti is much more challenging than the single object \ourdatareal data.
As expected, the best performing model (\opdformerp) for \ourdatamulti makes use of the per-part object pose prediction.  
There is a significant difference in performance between \opdformerp and \opdformero  for \ourdatamulti, but less for single objects.
This is because in the single object scenario having one global pose is sufficient.

\mypara{Analysis by number of openable objects.}
To better understand performance on \ourdatamulti we evaluate on all images in \ourdatamulti grouping into images with zero, one, or multiple openable (articulated) objects (AO).
\Cref{tab:results-OPDMulti-split} shows that \opdrcnn-based methods are better at avoiding false predictions on images without any openable parts.
For images with one or more openable objects, \opdformerp makes the most accurate predictions (highest +\mtype{}\axis{}\orig).
We also see that multiple AO is more challenging with both the part detection (\partdet) and motion parameter predictions (+\mtype{}\axis{}\orig) being much lower than the single AO case.

\mypara{What part types are more challenging?}
We find that \lid is the most challenging to detect on \ourdatacad.
We suspect this is due to limited data and variability of the \lid shape.
See the supplement for a detailed analysis.

\begin{figure*}[htbp]
\centering
\setkeys{Gin}{width=\linewidth}
\begin{tabularx}{0.9\linewidth}{Y Y Y Y Y Y}
\toprule

\small{GT} &
\imgclip{0}{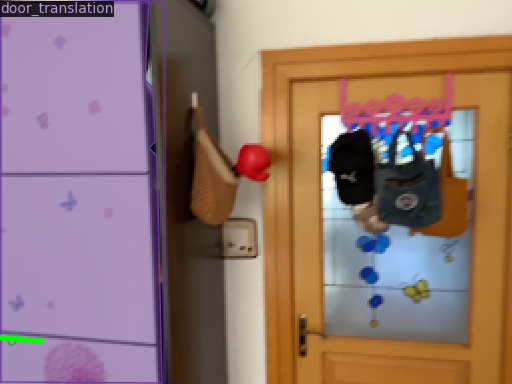} & 
\imgclip{0}{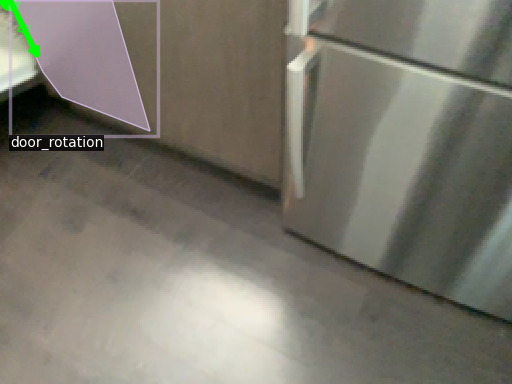} &
\imgclip{0}{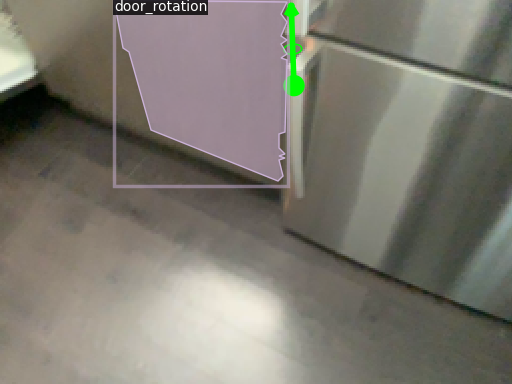} &
\imgclip{0}{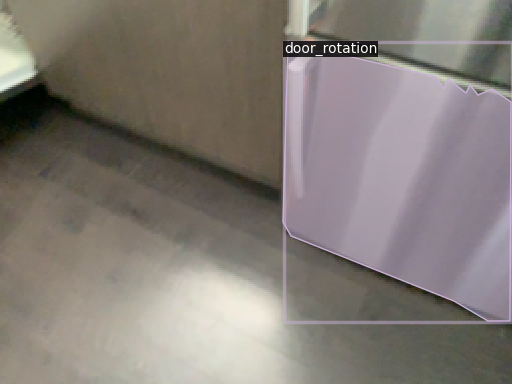} &
\imgclip{0}{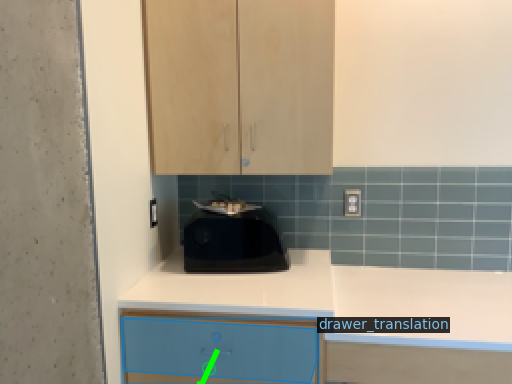}\\

\small{\opdrcnnop} &
Miss & 
Miss &
\imgclip{0}{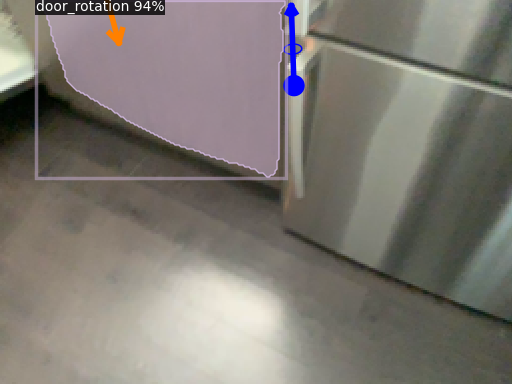} &
\imgclip{0}{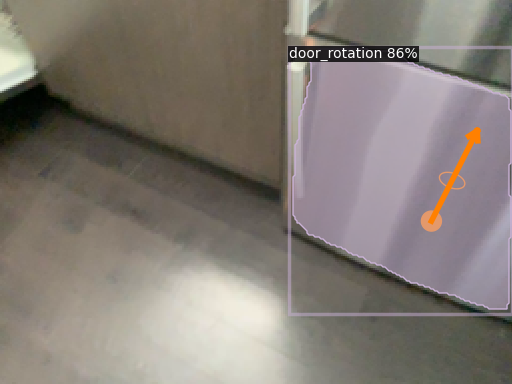} &
Miss \\
\small{\opdformerp} &
\imgclip{0}{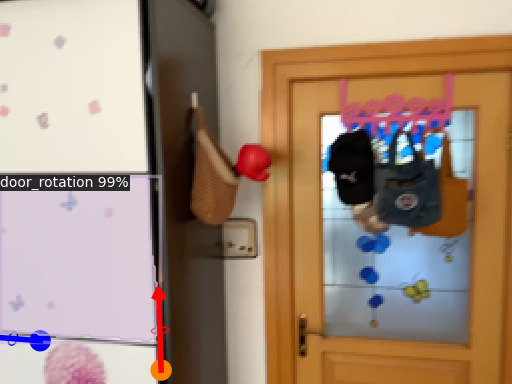} & 
Miss &
\imgclip{0}{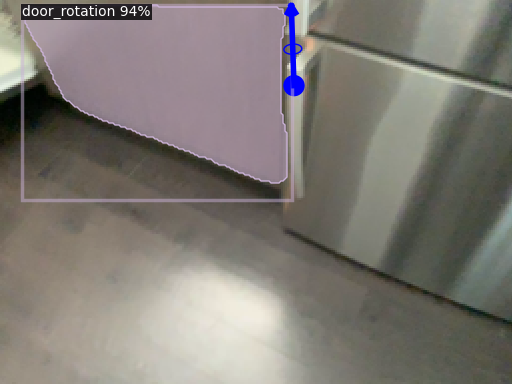} &
\imgclip{0}{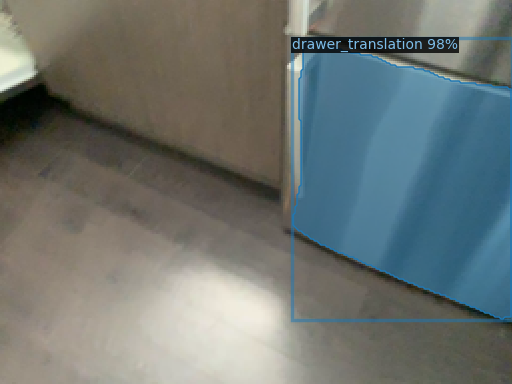} &
\imgclip{0}{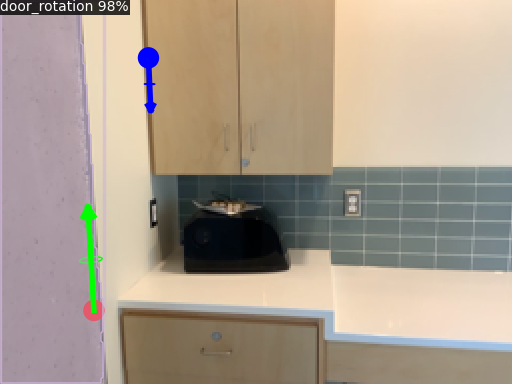}\\
\bottomrule
\end{tabularx}
\vspace{-5pt}
\caption{
Example failure cases.
Failures are due to limited camera field-of-view leading to cropping (1st column), unclear part edges (2nd, 3rd columns), confusion between door and drawer parts (4th column), and confusion between walls and doors.
}
\label{fig:fail}
\end{figure*}

\mypara{How good are the predicted object poses?}
To check whether \opdformerbaseline provides improved object pose predictions, we evaluate the object pose directly by measuring the rotation error (angle between two rotation matrices) and translation error (Euclidean distance normalized by the object diagonal length).
Following prior work~\cite{li2021leveraging}, we report the median error and accuracy at different thresholds.
For rotation accuracy, we use thresholds of $5^\circ$ degree, and for translation accuracy, we use a threshold of $0.1$.
\Cref{tab:results-pose-error-mini} shows the results of above object pose evaluation metric on the val sets of \ourdatacad and \ourdatamulti. We can see that \opdformerbaseline variants all  outperform \opdrcnno, indicating that the transformer structure can give better pose predictions.  For \ourdatamulti, the rotation and translation error are much higher than for \ourdatacad, illustrating the challenge of our \ourdatamulti scenario. 
Furthermore, our \opdformerp has better object pose prediction than \opdformero, indicating the importance of having per-part object pose prediction.  
In the single setting (\ourdatacad), the part-weight-average global pose gives the best object pose prediction.

\begin{table}
\centering
\resizebox{0.8\linewidth}{!}
{
\begin{tabular}{@{} ll rrrr @{}}
\toprule
& & \multicolumn{4}{c}{Part-averaged mAP $\% \uparrow$} \\
\cmidrule(l{0pt}r{2pt}){3-6}
Dataset & backbone & \partdet & +\mtype & +\mtype{}\axis & +\mtype{}\axis{}\orig \\
\midrule
\multirow{2}{*}{\ourdatacad}
& R50 & 79.0 & 76.7 & 58.6 & 53.4 \\
& Swin-L & \best{79.6} & \best{77.0} & \best{64.1} & \best{57.9} \\
\midrule
\multirow{2}{*}{\ourdatareal}
& R50 & 58.8 & 56.2 & 35.4 & 33.7\\
& Swin-L & \best{69.2} & \best{66.6} & \best{44.0} & \best{40.7} \\
\midrule
\multirow{2}{*}{\ourdatamulti}
& R50 & 32.9 & 31.6 & 19.4 & 16.0 \\
& Swin-L & \best{42.2} & \best{40.6} & \best{26.4} & \best{23.4} \\
\bottomrule
\end{tabular}
}
\vspace{-5pt}
\caption{Comparison of backbones with \opdformerp architecture on the val set for all three datasets.} 
\label{tab:results-backbone}
\end{table}
\mypara{Effect of backbone.}
Most of our experiments use the R50 backbone as it is smaller and requires fewer resources to train.
We check performance with Swin-L, a more powerful backbone compared to R50.
\Cref{tab:results-backbone} shows that with the Swin-L backbone, \opdformerp outperforms the R50 backbone in all cases.
Even when the part detection performance is roughly the same for OPDSynth dataset, the motion prediction is considerably higher (by 4.5\%).
Note that \opdformerp with Swin-L backbone (with 200 queries for the transformer decoder) has 205.6M parameters, which is around $5\times$ larger than the R50 backbone.

\mypara{Failure case analysis.}
\Cref{fig:fail} shows some failure cases.
Many errors occur due to the limited field-of-view and significant cropping of openable parts (see first column).
In the second and third column unclear edges lead to part detection failures.
In the fourth column motion type prediction fails due to a rotating door with drawer-like features.
The last column is an incorrect prediction of a wall as a door.

\section{Limitations}

Our work relies on projecting annotations from RGBD reconstructions in the MultiScan dataset to RGB frames.
Noise and errors in the reconstruction compound with potential annotation errors and can produce inaccurate projected 2D annotations for the openable part masks and motion parameters.
Moreover, the viewpoints from such RGBD video trajectories are biased by the path the human operators took to acquire a reconstruction and may not represent common viewpoints well.
The diversity of objects and scenes is also limited by geographic bias.
Furthermore, the sparsity of available real-world scene data with part-level motion annotations is a bottleneck for future work.

\section{Conclusion}

We generalized the openable part detection task to scenes with multiple objects.
To study this more realistic task setting, we constructed a dataset of images from real-world scenes and developed \opdformerbaseline, a part-informed transformer architecture that leverages insights about strong correlation between parts, and between object pose and parts.
We systematically evaluate on datasets from prior work and our new dataset to show that \opdformerbaseline achieves state-of-the-art performance on both single-object and multiple-object scenarios.
Our results show that scenarios with multiple openable objects in real scenes remain challenging, leaving opportunities for future work.
We hope our work catalyzes further investigation of openable part detection, enabling progress in 3D scene understanding, robotic vision, and embodied AI.

\mypara{Acknowledgements:}
This work was funded in part by a Canada CIFAR AI Chair, a Canada Research Chair and NSERC Discovery Grant, and enabled in part by support from \href{https://www.westgrid.ca/}{WestGrid} and \href{https://www.computecanada.ca/}{Compute Canada}.
We thank Yongsen Mao for helping us with the data processing procedure.  We also thank Jiayi Liu, Sonia Raychaudhuri, Ning Wang, Yiming Zhang for feedback on paper drafts.

{\small
\bibliographystyle{plainnat}
\setlength{\bibsep}{0pt}
\bibliography{main}
}

\clearpage
\newpage 
\appendix

In this supplement, we provide dataset details (\Cref{sec:supp:opdmulti}) and additional quantitative and qualitative results (\Cref{sec:supp:results}).

\section{Additional dataset details}
\label{sec:supp:opdmulti}

We provide more statistics for the \ourdatamulti dataset (\Cref{sec:supp-parts-dist}), examples of different part ratio coverages in frames (\Cref{sec:supp-part-coverage}), details about correcting inaccurate masks (\Cref{sec:supp-mask-correction}), and distribution of part and motion types for the train/val/test splits (\Cref{sec:supp-part-dist-expr}).  We also provide example visualization clarifying the difference between the different coordinate frames we consider for our models (\Cref{sec:supp-coord-frames}) and information about how we extract object poses for training (\Cref{sec:supp-object-pose}).

\subsection{Distribution of openable parts}
\label{sec:supp-parts-dist}

The \ourdatamulti dataset we constructed is composed of approximately $64K$ RGBD frames extracted from the MultiScan dataset~\cite{mao2022multiscan}.
Since our task focuses on detecting openable parts in images from more realistic real-world scenes with variable number of parts, we report the distribution of the openable parts across the dataset frames (see \Cref{tab:data-opdmulti-partstats}).
In \ourdatamulti, around $60\%$ of the frames that have at least one openable object (ignoring the frames that has 0 part) contain 1 part.
This distribution is approximately the same across train/val/test sets.
Overall, we observe that there is a long-tail distribution for the number of openable parts observed in these RGBD frames captured by people from real scenes.

\begin{table}
\centering
\begin{tabular}{@{} l rrrrr @{}}
\toprule
& \multicolumn{5}{c}{\# Parts}\\
\cmidrule(lr{0pt}){2-6}
split & 0 & 1 & 2 & 3 & 4+
\\
\midrule
train & 31189 & 7705 & 3183 & 1261 & 664 \\
val & 6424 & 1853 & 1055 & 516 & 320 \\
test & 6818 & 1839 & 826 & 336 & 224 \\
\midrule
total & 44431 & 11397 & 5064 & 2113 & 1208 \\
\bottomrule
\end{tabular}
\caption{Distribution of openable parts over train/val/test splits, indicating the number of frames having different numbers of openable parts.}
\label{tab:data-opdmulti-partstats}
\end{table}

\subsection{Part mask image coverage ratio examples}
\label{sec:supp-part-coverage}

\begin{figure}
\centering
\setkeys{Gin}{width=\linewidth}
\begin{tabularx}{\linewidth}{ Y Y Y Y}

\toprule

$(0-5] \%$ &
$(5-10] \%$ &
$(10-15] \%$ &
$\ge15 \%$ \\

\midrule
\imgclip{0}{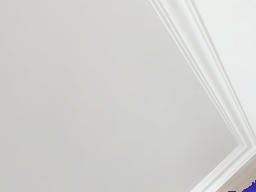} &
\imgclip{0}{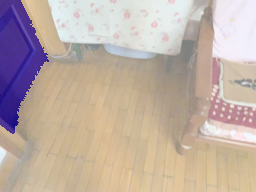} &
\imgclip{0}{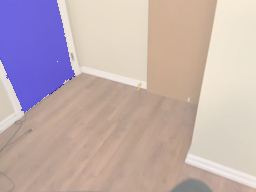} &
\imgclip{0}{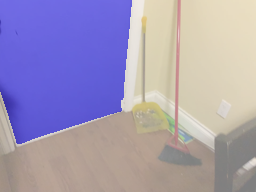}\\

\imgclip{0}{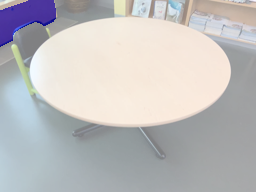} &
\imgclip{0}{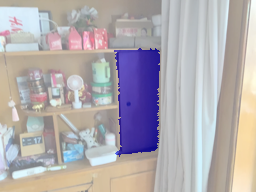} &
\imgclip{0}{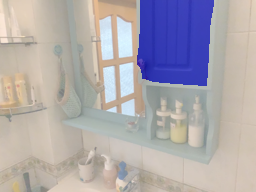} &
\imgclip{0}{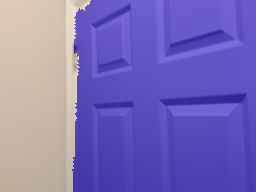}\\

\bottomrule
\end{tabularx}
\caption{
Examples of \ourdatamulti frames with different per-part pixel ratio (percent of pixels in that frame for a given part).  We note that when the pixel ratio for a part is extremely low (<5\%), it is challenging for humans to recognize the part.  Thus, we do not include such parts in our evaluation.  
}
\label{fig:pix-ratio}
\end{figure}

We show examples of parts with different part coverage ratios (e.g. fraction of the frame covered by the part) in \Cref{fig:pix-ratio}. 
We note that images with low part coverage ratio (below 5\%) do not provide sufficient information for openable part detection, thus we exclude these parts from our evaluation.

\begin{figure}[t]
\includegraphics[width=1\linewidth]{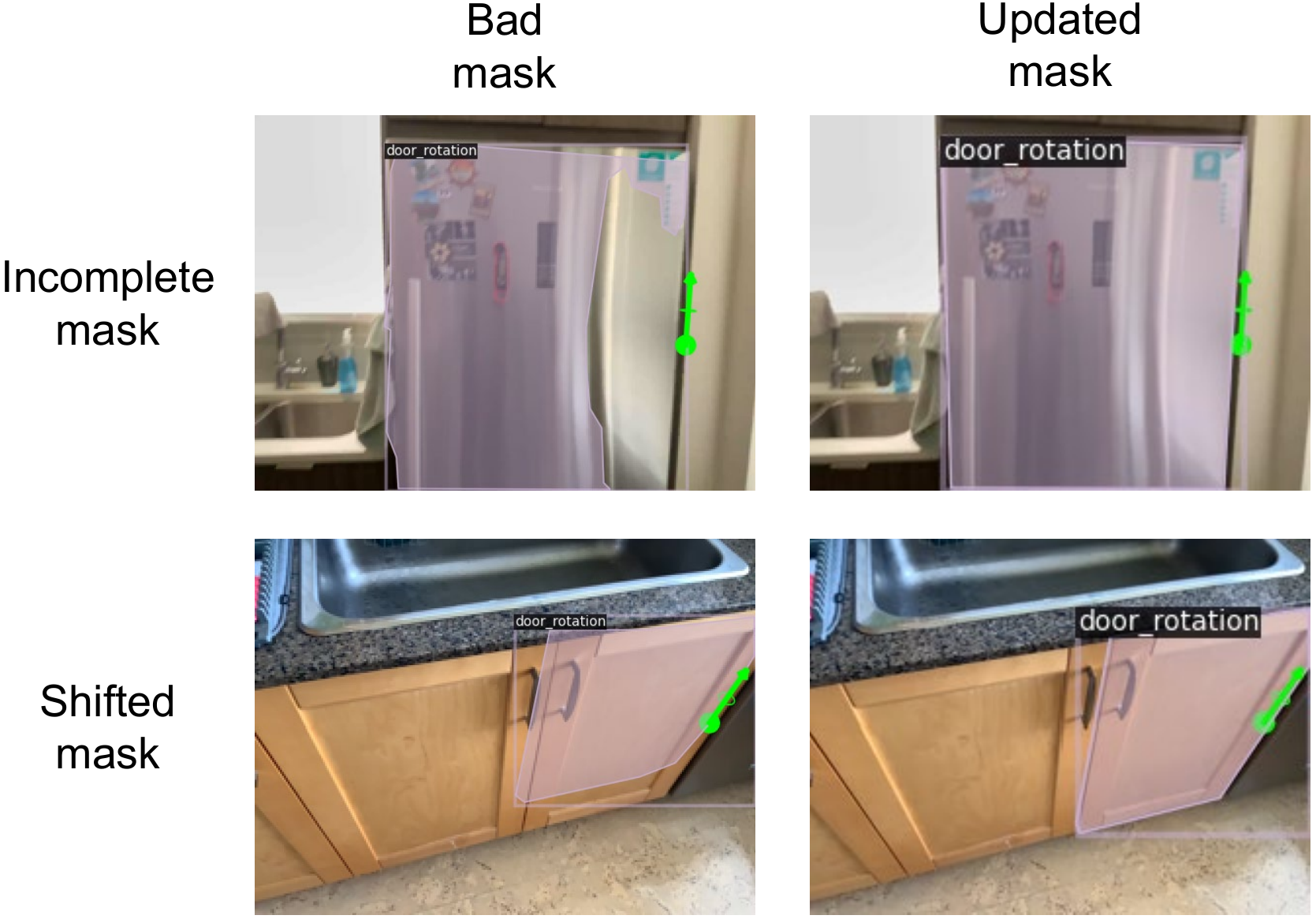}
\caption{Examples of inaccurate openable part masks and corresponding updated masks. The left column shows the masks with errors, while the right column shows the updated maskes. The first row is an example of the incomplete mask error, and the second row shows the shifted mask error.
}
\label{fig:mask_correct}
\end{figure}
\subsection{Mask correction}
\label{sec:supp-mask-correction}

As noted in the main paper, it is possible for some frames to have inaccurate masks due to issues in the reconstruction.  Since the annotations are on the 3D reconstructions, if the reconstruction is incomplete or the camera pose for the frame is inaccurate, then the annotated mask when projected onto the image will be inaccurate.
\Cref{fig:mask_correct} shows examples of inaccurate projected mask annotations.  From the top row, we see an example of an incomplete mask due to a shiny surface that is not reconstructed well.
The bottom row shows an example of a shifted mask due to an inaccurate camera pose estimate for the frame.
We note that despite the inaccurate masks, the projected motion axis is good.  

To ensure that the data for our evaluation is accurate and high-quality, we manually inspect all frames in the validation and test splits and flag frames with bad mask annotations.
We then use the Toronto Annotation Suite \cite{torontoannotsuite} to re-annotate the bad masks manually.
Considering there may be multiple parts in one frame, we directly re-annotate on the images with bad part masks.
During the annotation procedure, we draw the polygon segmentation for the specific openable part.
In total, we re-annotate 2261 openable part mask segmentations across both the validation and test sets.

\begin{table}[t]
\resizebox{\linewidth}{!}{
\centering
{
\begin{tabular}{@{} l r rrr rr@{}}
\toprule
& & \multicolumn{3}{c}{Part Type} & \multicolumn{2}{c}{Motion Type}\\
\cmidrule(l{0pt}r{2pt}){3-5} \cmidrule(l{2pt}r{2pt}){6-7} 
split & \# part & \door & \drawer & \lid & \mtrot & \mttrans\\
\midrule
train & 18479 & 15904 & 2097 & 478 & 15760 & 2719 \\
val & 6077 & 5124 & 752 & 201 & 4695 & 1382\\
test & 4704 & 3592 & 981 & 131 & 3449 & 1255 \\
\midrule
total & 29260 & 24620 & 3830 & 810 & 23904 & 5356 \\
\bottomrule
\end{tabular}
}}
\caption{OPDMulti statistics for train/val/test splits for part type and motion type.  These statistics are computed after discarding small parts that occupy less than 5\% of a frame (see \Cref{sec:supp-part-coverage}).  Most of the parts we observe in this real world dataset is revolute doors.
}
\label{tab:stats-motion-part}
\end{table}

\subsection{Distribution of motion and openable part type}
\label{sec:supp-part-dist-expr}

We report the distribution of different motion types and openable part types in the \ourdatamulti dataset in \Cref{tab:stats-motion-part}.
This distribution is counted from the data we used for our experiments, which means it is after excluding small parts as described in \Cref{sec:supp-part-coverage}.
From \Cref{tab:stats-motion-part}, we see that \door is the most frequent part type with around $83\%$ ratio in all splits, and \mtrot is more common than \mttrans.
Since \ourdatamulti is constructed from real-world indoor scenes, the distribution shows the frequency of openable parts in a realistic setting.
We analyze the model performance for different part types in \Cref{sec:supp-analysis}.

\subsection{Coordinate frames}
\label{sec:supp-coord-frames}

\begin{figure}[t]
\includegraphics[width=1\linewidth]{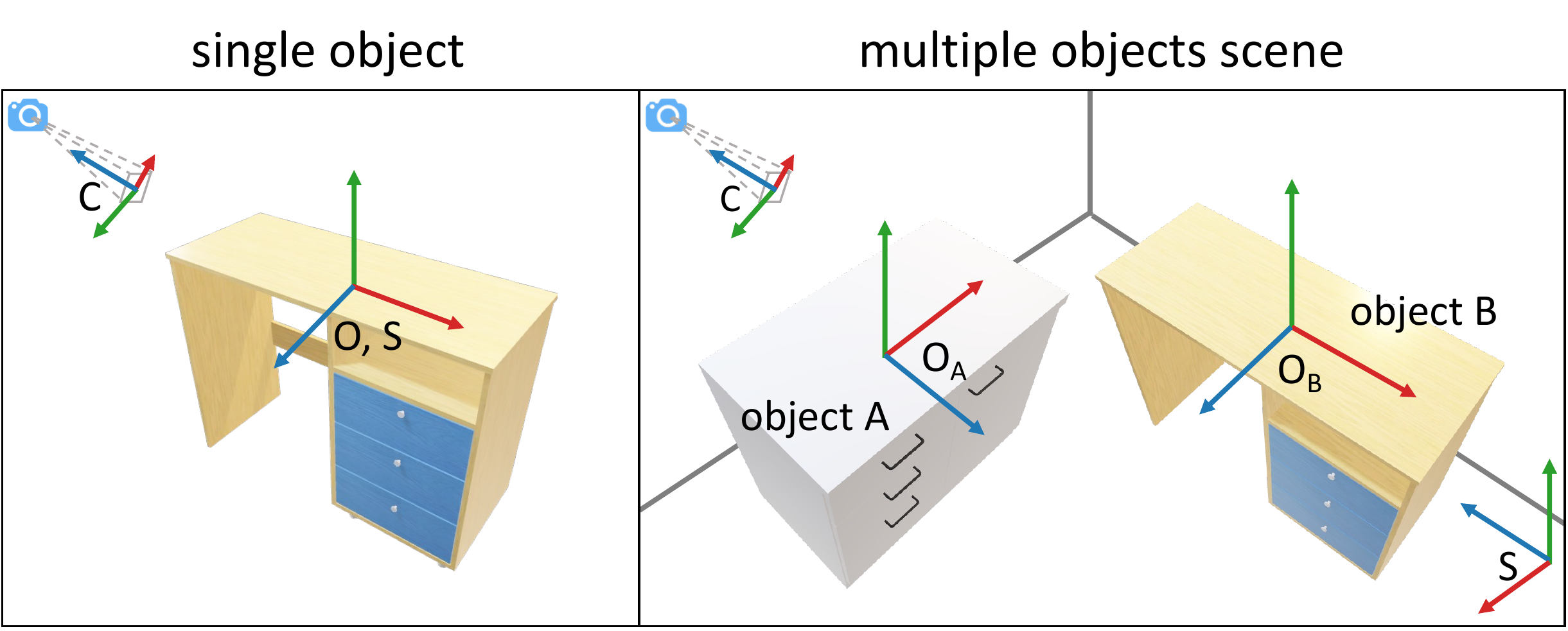}
\caption{We show the difference between the coordinate frames we use in our work.   In the case of a single object (left), the global scene coordinate (\textbf{S}) is the same as the single object coordinate (\textbf{O}).  When there are multiple objects in the scene, we need to predict multiple object poses as each object has its own coordinate frame ($\mathbf{O_A}$ and $\mathbf{O_B}$ for objects $\mathbf{A}$ and $\mathbf{B}$).
All motion parameters are evaluated in camera coordinates (\textbf{C}).
We use the colors RGB to represent the local coordinate system XYZ axes in that order.
}
\label{fig:coor}
\end{figure}

\Cref{fig:coor} illustrates the difference between the different coordinate frames.
\citet{jiang2022opd}'s dataset contained images with only a single object (see \Cref{fig:coor} left).
When a scene contains only one object, the object coordinate is the same as the global scene coordinate.
In contrast, when there are multiple objects, the global scene coordinates and the object coordinate frame can be different.
The object coordinate frame ($\mathbf{O_A}$ and $\mathbf{O_B}$ for objects $\mathbf{A}$ and $\mathbf{B}$ in \Cref{fig:coor}) for each object are centered at the center of each object, with the axes aligned to the bounding box of the object (with +Z as front, and +Y as up), while the global scene coordinate is a single coordinate frame at the center of the scene.
The camera coordinate frame is used for projecting the 3D scene onto the image plane and has +Z facing into the camera, and +Y as up. 

\begin{figure}
\centering
\setkeys{Gin}{width=\linewidth}
\begin{tabularx}{\linewidth}{@{} Y @{\hskip 5pt} Y @{}}
Scene & Objects\\
\imgclip{5pt}{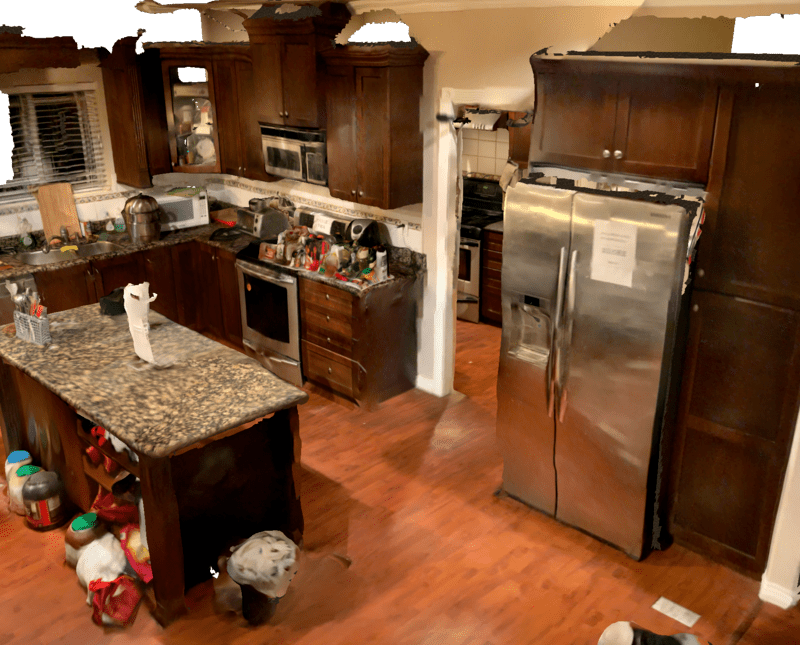} &
\imgclip{5pt}{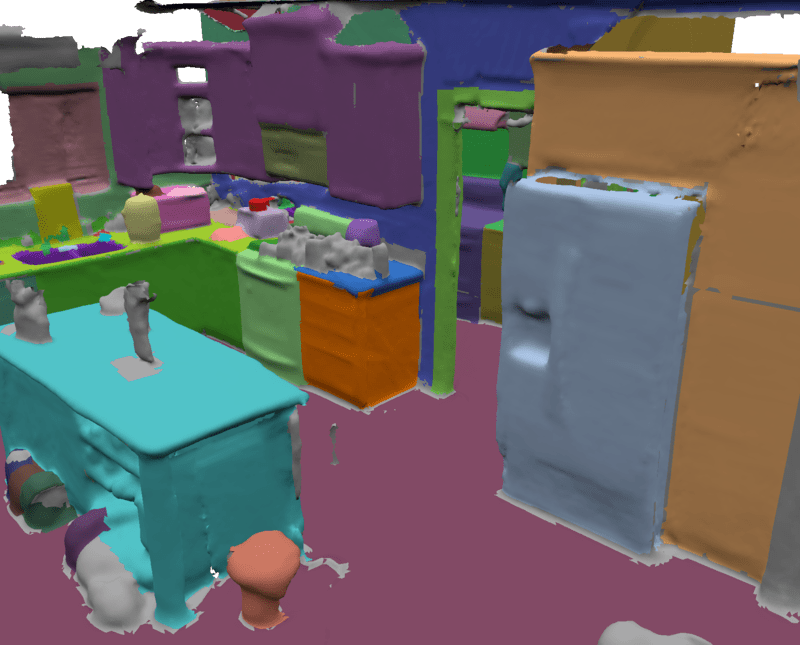}\\
Parts & Object poses\\
\imgclip{5pt}{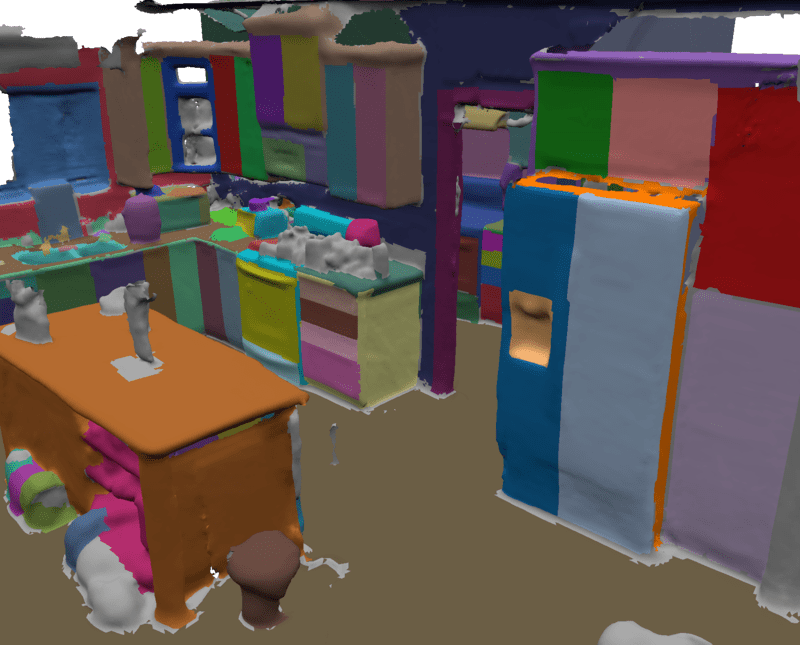} &
\imgclip{5pt}{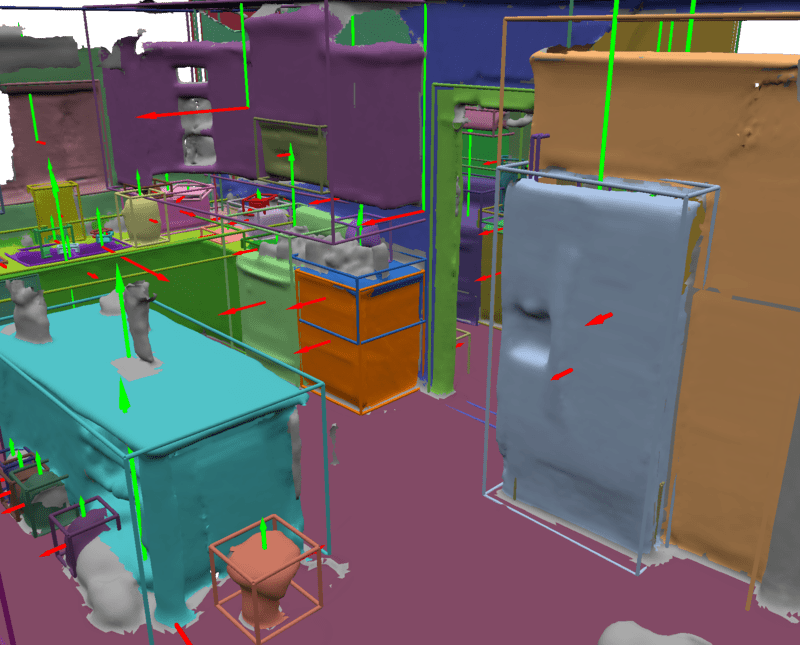}\\
\end{tabularx}
\caption{
Example scene used in our \ourdatamulti dataset.
The scene contains several objects (colored in top right), each with potentially multiple openable parts (colored in bottom left).
The object poses are indicated by axes showing the up (green) and front (red) for each object. 
}
\label{fig:pose_vis}
\end{figure}

\subsection{Object pose for training}
\label{sec:supp-object-pose}

\Cref{fig:pose_vis} shows an example kitchen scene in our dataset containing multiple objects with openable parts.
In such scenarios, some care has to be taken to construct appropriate single global pose and multiple object poses information to train each of the architecture variants.

\mypara{Single global pose}.
Since \ourdatamulti images contain an arbitrary number of objects, there is no notion of a single global object pose.
Therefore, we train with the scene coordinates defining a global pose, and transform from camera coordinates to these scene coordinates.
All relevant motion parameters are also transformed to the scene coordinates.

\mypara{Multiple object poses}.
In this case, the object pose is defined by the transformation between camera coordinates and canonical object coordinates, using a consistently defined front and up orientation for each object pose.
See \Cref{fig:pose_vis} bottom right for example object coordinate frames.

\section{Additional results}
\label{sec:supp:results}

In this section, we provide information about the size of our models (\Cref{sec:supp-model-size}), additional quantitative results on depth, RGB-D images, and the test set (\Cref{sec:supp-quant-results}), breakdown of \opdformerp performace on different parts and comparisons of different training strategies for \opdformerp (\Cref{sec:supp-analysis}).  In addition, we provide additional qualitative examples (\Cref{sec:supp-qual-results}), visualization of the transformer attention maps (\Cref{sec:supp-attention}), and a discussion of object part consistency (\Cref{sec:supp-part-consistency}).

\begin{table}
\centering
{
\begin{tabular}{@{} l ccc @{}}
\toprule
Model & Backbone & Parameters & FLOP
\\
\midrule
\opdrcnnc & R50 & 44.5M & 47.4G \\
\opdrcnno & R50 & 45.1M & 49.3G \\
\opdrcnnop & R50 & 46.1M & 46.9G \\
\opdformerc & R50 & 41.9M & 20.3G\\
\opdformero & R50 & 43.5M & 20.3G \\
\opdformerp & R50 & 42.0M & 20.3G\\
\opdformerp & Swin-L & 205.6M & 92.1G\\
\bottomrule
\end{tabular}
}
\caption{
Parameter and FLOP counts for models with R-50 and Swin-L backbone.
\opdrcnn based methods have more parameters than \opdformerbaseline based models.
The Swin-L backbone is 5 times larger than the R50 backbone
}
\label{tab:model-param-count}
\end{table}

\begin{table}
\resizebox{\linewidth}{!}
{
\begin{tabular}{@{} ll rrrr @{}}
\toprule
& & \multicolumn{4}{c}{Part-averaged mAP $\% \uparrow$} \\
\cmidrule(l{0pt}r{2pt}){3-6}
Dataset & Model & \partdet & +\mtype & +\mtype{}\axis & +\mtype{}\axis{}\orig \\
\midrule
\multirow{6}{*}{\ourdatacad}
& \opdrcnnc~\cite{jiang2022opd} & 69.5\stdhide{0.32} &	67.7\stdhide{0.36} &	37.7\stdhide{0.27} &	35.3\stdhide{0.24} \\
& \opdrcnno~\cite{jiang2022opd} & 69.3\stdhide{0.34} &	67.5\stdhide{0.33} &	50.7\stdhide{0.58} &	44.8\stdhide{0.45} \\
& \opdrcnnop & 68.6\stdhide{0.71} &	66.8\stdhide{0.69} &	48.7\stdhide{0.56} &	42.5\stdhide{0.56} \\
& \opdformerc  & 78.9 & 76.9 & 51.2 & 47.8 \\
& \opdformero & \best{79.6\stdhide{0.26}} & \best{78.1\stdhide{0.24}} & \best{62.4\stdhide{0.52}} & \best{54.9\stdhide{0.61}} \\
& \opdformerp & 77.6 & 75.8 & {56.6} & {53.2} \\

\midrule
\multirow{6}{*}{\ourdatareal}
& \opdrcnnc~\cite{jiang2022opd} & 40.0 & 38.0 & 7.2 & 6.7 \\
& \opdrcnno~\cite{jiang2022opd} & 39.9 & 37.8 & 12.6 & 11.1 \\
& \opdrcnnop &  40.0 & 37.6 & 18.8 & 17.9 \\
& \opdformerc & 48.3 & 47.4 & 29.9 & 28.8 \\
& \opdformero & {50.4\stdhide{0.22}} & {49.6\stdhide{0.15}} & 32.4\stdhide{0.19} & 30.3\stdhide{0.26} \\
& \opdformerp & \best{50.8} & \best{50.1} & \best{37.3} & \best{35.1}\\

\midrule
\multirow{6}{*}{\ourdatamulti}
& \opdrcnnc~\cite{jiang2022opd} & 18.9 & 16.5 & 2.5 & 2.3 \\
& \opdrcnno~\cite{jiang2022opd} & 17.3 & 15.1 & 0.8 & 0.1 \\
& \opdrcnnop & 18.8 & 16.3 & 4.4 & 3.1 \\
& \opdformerc & 22.1 & 19.9 & 11.4 & 10.2 \\
& \opdformero & \best{24.9} & \best{22.6} & 5.8 & 1.9 \\
& \opdformerp & 23.0 & 20.8 & \best{16.1} & \best{13.9} \\
\bottomrule
\end{tabular}
}
\caption{Comparison of \opdrcnn and \opdformerbaseline on validation set \textbf{depth} input images for the three datasets (\ourdatacad and \ourdatareal for single objects, and \ourdatamulti for multiple-object real scenes).}
\label{tab:results-OPDAll-val-depth-mini}
\end{table}
\begin{table}
\resizebox{\linewidth}{!}
{
\begin{tabular}{@{} ll rrrr @{}}
\toprule
& & \multicolumn{4}{c}{Part-averaged mAP $\% \uparrow$} \\
\cmidrule(l{0pt}r{2pt}){3-6}
Dataset & Model & \partdet & +\mtype & +\mtype{}\axis & +\mtype{}\axis{}\orig \\
\midrule
\multirow{6}{*}{\ourdatacad}
& \opdrcnnc~\cite{jiang2022opd} & 72.8\stdhide{0.67} &	70.6\stdhide{0.68} &	39.2\stdhide{0.40} &	36.6\stdhide{0.41} \\
& \opdrcnno~\cite{jiang2022opd} & 72.5\stdhide{0.34} &	70.6\stdhide{0.29} &	51.7\stdhide{0.66} &	47.0\stdhide{0.62} \\
& \opdrcnnop & 70.5\stdhide{0.62} &	68.4\stdhide{0.62} &	49.6\stdhide{0.42} &	44.1\stdhide{0.44} \\
& \opdformerc & 77.1 & 75.0 & 49.6 & 45.8 \\
& \opdformero & \best{77.2\stdhide{0.38}} & 75.2\stdhide{0.43} & \best{60.8\stdhide{0.95}} & 54.0\stdhide{0.64} \\
& \opdformerp & \best{77.2 }& \best{75.3} & {59.7} & \best{55.7} \\

\midrule
\multirow{6}{*}{\ourdatareal}
& \opdrcnnc~\cite{jiang2022opd} & 56.2 & 54.1 & 15.1 & 14.6 \\
& \opdrcnno~\cite{jiang2022opd} & 55.8 & 53.3 & 30.0 & 27.5 \\
& \opdrcnnop & 57.8 & 54.7 & 30.2 & 28.0 \\
& \opdformerc & \best{65.0} & \best{62.1} & 34.4 & 33.7 \\
& \opdformero & 61.6\stdhide{0.27} & 60.1\stdhide{0.29} & 36.3\stdhide{0.35} & 34.1\stdhide{0.22} \\
& \opdformerp & {61.6} & {58.9} & \best{40.7} & \best{39.7} \\

\midrule
\multirow{6}{*}{\ourdatamulti}
& \opdrcnnc~\cite{jiang2022opd} & 23.4 & 21.1 & 6.8 & 6.0 \\
& \opdrcnno~\cite{jiang2022opd} & 23.2 & 21.2 & 2.9 & 0.6 \\
& \opdrcnnop & 25.5 & 23.6 & 9.1 & 7.8 \\
& \opdformerc & 25.3 & 23.6 & 14.2 & 13.5 \\
& \opdformero & 24.1 & 22.0 & 6.6 & 2.6 \\
& \opdformerp & \best{28.6} & \best{26.5} & \best{18.7} & \best{17.2} \\
\bottomrule
\end{tabular}
}
\caption{Comparison of \opdrcnn and \opdformerbaseline on validation set \textbf{RGBD} input images for the three datasets (\ourdatacad and \ourdatareal for single-object, and \ourdatamulti for multiple-object real scenes).}
\label{tab:results-OPDAll-val-rgbd-mini}
\end{table}

\begin{table}
\resizebox{\linewidth}{!}{
\begin{tabular}{@{} ll rrrr rrr @{}}
\toprule
& & \multicolumn{4}{c}{Part-averaged mAP $\% \uparrow$} \\
\cmidrule(l{0pt}r{2pt}){3-6} 
Dataset & Model & \partdet & +\mtype & +\mtype{}\axis & +\mtype{}\axis{}\orig \\
\midrule
\multirow{6}{*}{\ourdatacad}
& \opdrcnnc & 67.4\stdhide{0.26} & 66.2\stdhide{0.18} & 40.9\stdhide{0.21} & 38.0\stdhide{0.19} \\
& \opdrcnno & 66.6\stdhide{0.28} & 65.5\stdhide{0.27} & 50.8\stdhide{0.20} & 47.0\stdhide{0.25} \\
& \opdrcnnop & 66.7 & 65.1 & 49.9 & 45.8 \\
& \opdformerc  & 69.0 & 67.7 & 52.4 & 49.0\\
& \opdformero & 68.6\stdhide{0.78} & 67.4\stdhide{0.45} & \best{56.3\stdhide{0.32}} & \best{53.2\stdhide{0.40}}\\
& \opdformerp & \best{72.8} & \best{71.2} & 55.9 & 52.1 \\
\midrule
\multirow{6}{*}{\ourdatareal}
& \opdrcnnc & \best{58.0} & \best{57.0} & 22.2 & 21.3 \\
& \opdrcnno & 57.8 & 56.4 & 33.1 & 30.7  \\
& \opdrcnnop & 55.9 & 52.2 & 33.0 & 31.3  \\
& \opdformerc &  54.7 & 50.6 & 34.1 & 32.9 \\
& \opdformero & 54.4\stdhide{1.42} & 51.9\stdhide{1.91} & 35.8\stdhide{0.95} & 33.6\stdhide{0.87} \\
& \opdformerp  & 57.3 & 55.6 & \best{39.7} & \best{38.1} \\
\midrule
\multirow{6}{*}{\ourdatamulti}
& \opdrcnnc & 25.2 & 24.7 & 5.6 & 4.7 \\
& \opdrcnno & 16.9 & 16.1 & 3.6 & 0.8 \\
& \opdrcnnop & 22.6 & 21.7 & 5.5 & 4.2 \\
& \opdformerc & 29.8 & 29 & 9.3 & 8.2 \\
& \opdformero & 26.9 & 25.3 & 3.8 & 1.4 \\
& \opdformerp & \best{31.7} & \best{30.7} & \best{18.4} & \best{16.8 }\\
\bottomrule
\end{tabular}
}
\caption{Experiment results for different methods on the \textbf{test} set of \ourdatacad, \ourdatareal, and \ourdatamulti for \textbf{RGB} input. Best performing method in bold, and second best in bold and italics. Results on the test set follow the same trends as the val set, with \opdformerbaseline methods outperforming \opdrcnn counterparts.}
\label{tab:results-OPDAll-test}
\end{table}

\begin{table*}[t]
\resizebox{\linewidth}{!}{
\centering
{
\begin{tabular}{@{} ll rrrr rrrr rrrr @{}}
\toprule
& & \multicolumn{4}{c}{\drawer} & \multicolumn{4}{c}{\door} & \multicolumn{4}{c}{\lid} \\
\cmidrule(l{0pt}r{2pt}){3-6} \cmidrule(l{2pt}r{2pt}){7-10} \cmidrule(l{2pt}r{0pt}){11-14}
Dataset & Model & \partdet & +\mtype & +\mtype{}\axis & +\mtype{}\axis{}\orig & \partdet & +\mtype & +\mtype{}\axis & +\mtype{}\axis{}\orig & \partdet & +\mtype & +\mtype{}\axis & +\mtype{}\axis{}\orig \\
\midrule
\multirow{4}{*}{\ourdatacad} 
& \opdrcnnc~\cite{jiang2022opd} & 81.3 & 80.9 & 60.7 & 60.7 & \best{85.3} & \best{79.8} & 46.2 & 42.8 & 57.5 & 57.3 & 13.5 & 5.7 \\
& \opdrcnnop & 79.8 & 79.6 & 68.8 & 68.8 & 85.1 & 80.1 & 59.0 & 53.0 & 57.1 & 56.9 & 27.6 & 12.9 \\
& \opdformerc & 81.4 & 80.8 & 68.8 & 68.8 & 82.9 & 76.4 & 54.6 & 50.3 & 67.6 & 67.5 & 23.3 & 12.5 \\
& \opdformerp & \best{83.1} & \best{82.3} & \best{73.5} & \best{73.5} & 84.0 & 78.0 & \best{60.0} & \best{55.8} & \best{68.3} & \best{68.1} & \best{39.9} & \best{27.4} \\
\midrule
\multirow{4}{*}{\ourdatareal} 
& \opdrcnnc &  77.6 & 76.8 & 23.0 & 23.0 & 56.0 & 51.7 & 22.5 & 20.3 & 39.4 & 26.1 & 0.8 & 0.1 \\
& \opdrcnnop & \best{76.9} & \best{76.3} & 48.3 & 48.3 & 57.9 & 52.3 & 30.2 & 26.8 & 38.2 & 36.2 & 2.4 & 0.2 \\
& \opdformerc & 74.6 & 73.6 & 50.0 & 50.0 & 54.8 & 51.3 & 36.0 & 33.2 & \best{44.4} & \best{43.3} & \best{3.1} & \best{1.6} \\
& \opdformerp & 76.0 & 75.0 & \best{61.3} & \best{61.3} & \best{58.1} & \best{53.8} & \best{42.8} & \best{38.6} & 39.5 & 36.7 & 1.4 & 0.1 \\
\midrule
\multirow{4}{*}{\ourdatamulti} 
& \opdrcnnc & \best{24.3} & \best{24.1} & 8.1 & 8.1 & 41.9 & 37.4 & 16.1 & 13.5 & 15.6 & 15.6 & 2.2 & 1.8 \\
& \opdrcnnop & 13.9 & 13.5 & 4.5 & 4.5 & 39.1 & 33.8 & 15.1 & 10.7 & 9.6 & 9.5 & 2.0 & 1.9 \\
& \opdformerc & 21.1 & 20.6 & 9.8 & 9.8 & 43.3 & 39.7 & 26.4 & 23.6 & 26.4 & 26.4 & 3.1 & 2.9 \\
& \opdformerp & 22.3 & 21.5 & \best{15.3} & \best{15.3} & \best{44.2} & \best{41.1} & \best{27.9} & \best{23.5} & \best{32.2} & \best{32.1} & \best{15.1} & \best{9.0} \\
\bottomrule
\end{tabular}
}}
\caption{Breakdown of performance by part category, evaluated on RGB inputs. From the results, we see that \lid is the most challenging part.
From \Cref{tab:stats-motion-part}, we saw that in \ourdatamulti, the ratio of part types is $84\%/ 13\% /3\%$ for \door/\drawer/\lid respectively.
The amount of training data available for the different parts is one of the reasons that our model performs best for \door  and worst for \lid.
We also see that \opdformerp tend to have better motion prediction, even when the detection performance is not as good.
}
\label{tab:analysis-part-cat}
\end{table*}

\subsection{Model size}
\label{sec:supp-model-size}

In \Cref{tab:model-param-count}, we show the number of parameters and FLOPs for each of our models.  We see from \Cref{tab:model-param-count} that our \opdformerbaseline models have slightly less parameters than the \opdrcnn models.  The \opdformerp model with Swin-L backbone is considerably larger (5x). 

\subsection{Additional quantitative results}
\label{sec:supp-quant-results}

We present additional quantitative results comparing our proposed \opdformerbaseline with \opdrcnn, including results on depth and RGBD inputs.
We also present results on the test set.
Results consistently show that \opdformerp outperforms other variants. 

\mypara{Results for depth and RGBD inputs.}
\Cref{tab:results-OPDAll-val-depth-mini,tab:results-OPDAll-val-rgbd-mini} report the results using depth and RGBD as input with val set from three different datasets.
We see that the \opdformerbaseline variants perform better than \opdrcnn methods across input image formats.
Compared with using RGB input (see main paper Tab. 3), for part motion prediction, using D input only gives slightly worse results while using RGBD input gives slightly better results. 

\mypara{Results on the test set.}
\Cref{tab:results-OPDAll-test} evaluates the different methods on RGB images from the test set of the three datasets.  The performance on the test set largely follows that of the validation sets, with \opdformerbaseline variants outperforming \opdrcnn and per-part object pose (\opdformerp) providing the best performance.

\begin{table}
\centering
\resizebox{\linewidth}{!}
{
\begin{tabular}{@{} lll rrrr @{}}
\toprule
& & & \multicolumn{4}{c}{Part-averaged mAP $\% \uparrow$} \\
\cmidrule(l{0pt}r{2pt}){4-7}
&  Model & Initialized with & \partdet & +\mtype & +\mtype{}\axis & +\mtype{}\axis{}\orig \\
\midrule
1 & Mask2Former & full pretrained model on \ourdatareal & 32.8\std{0.73} & - & - & - \\
2 & \opdformerp & pretrained detection model on COCO & 28.1\std{0.67} & 26.6\std{0.59} & 12.3\std{0.43} & 10.7\std{0.29} \\
3 & \opdformerp & pretrained detection model on \ourdatamulti & 32.9\std{0.35} & 31.2\std{0.41} & 14.6\std{0.22} & 13.0\std{0.18} \\
4 & \opdformerp & full pretrained model on \ourdatareal & \best{32.9}\std{0.69} & \best{31.6}\std{0.72} & \best{19.4}\std{0.38} & \best{16.0}\std{0.03} \\

\bottomrule
\end{tabular}
}
\caption{Comparison between Mask2Former (row 1) with no extra losses for \opdformerp (initial weights are pretrained on OPDReal),  \opdformerp initialized with Mask2Former weights (row 2,3)  trained on just COCO or OPDMulti, and \opdformerp (row 4) with full weights that are pretrained from OPDReal.} 
\label{tab:results-mask2former-comp}
\end{table}

\subsection{Additional analysis}
\label{sec:supp-analysis}

\mypara{What part types are more challenging?}
\Cref{tab:analysis-part-cat} analyzes performance on three part categories: \drawer, \door, \lid.
We find \lid to be the most challenging to detect on all three datasets.
On \ourdatamulti, all three part types are much more challenging with considerably lower \partdet.

\mypara{Comparison between Mask2Former and \opdformerbaseline}. 
We compare the performance of our \opdformerbaseline with the original Mask2Former, without any additional motion losses, to check whether having extra losses would impact the detection performance. 
We report results on \ourdatamulti val set with RGB input in \Cref{tab:results-mask2former-comp}.  We start with the Mask2Former weights pretrained on the COCO dataset~\cite{lin2014microsoft} and then consider different training strategies. For all experiments except for (2), we pretrain on OPDReal before we train on \ourdatamulti. For Mask2Former, we initialize the model with weights pretrained on OPDReal.  For training \opdformerbaseline (row 1), we compare training from pretrained model just on COCO dataset (and not further pretrained on OPDReal, row 2), vs starting with the weights from the Mask2Former model pretrained on \ourdatamulti (row 3) vs training \opdformerbaseline directly from a pretrained model on \ourdatareal (row 4). 
From \Cref{tab:results-mask2former-comp}, we see that \opdformerbaseline performance without pretraining (row 2) on \ourdatareal is lower than models with pretraining (row 1,3,4).  
For models with pretraining, they have very similar part detection performance (\partdet) but pretraining the motion parameters on \ourdatareal (row 4) results in better motion prediction, especially for motion axis and origin (+\mtype{}\axis, +\mtype{}\axis{}\orig).  

\subsection{Additional qualitative results}
\label{sec:supp-qual-results}

\begin{figure*}
\centering
\setkeys{Gin}{width=\linewidth}
\begin{tabularx}{\textwidth}{@{} Y Y Y Y Y Y @{}}

\toprule


\textbf{\ourdatacad} \\
\midrule

\small{GT} &
\imgclip{0}{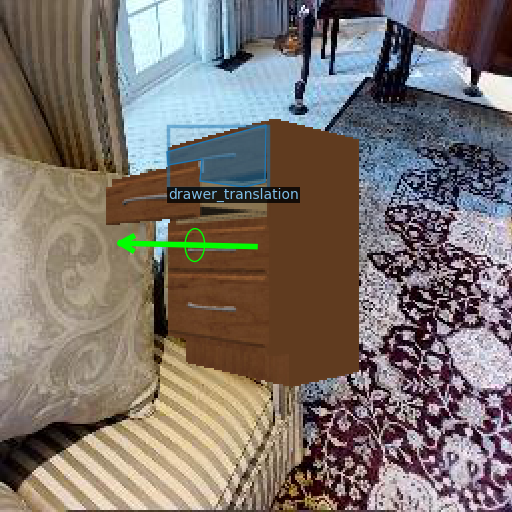} & 
\imgclip{0}{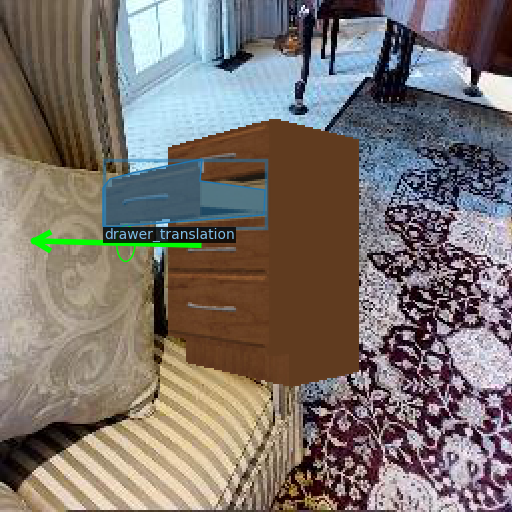} &
\imgclip{0}{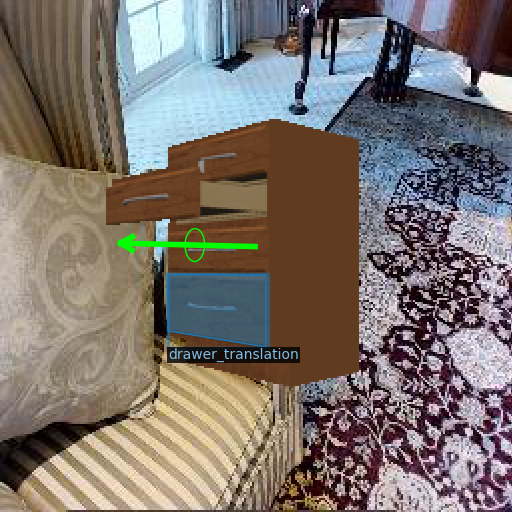} &
\imgclip{0}{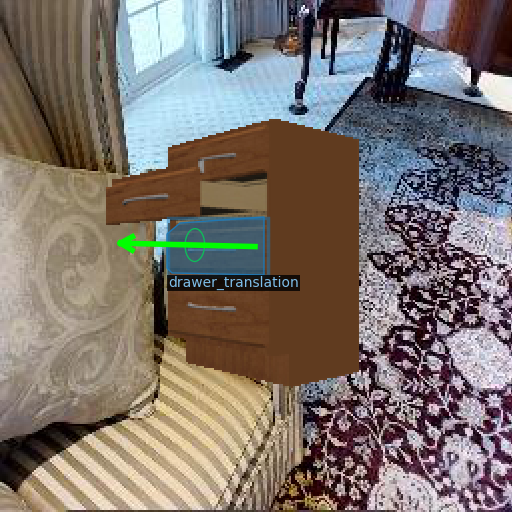} &
\imgclip{0}{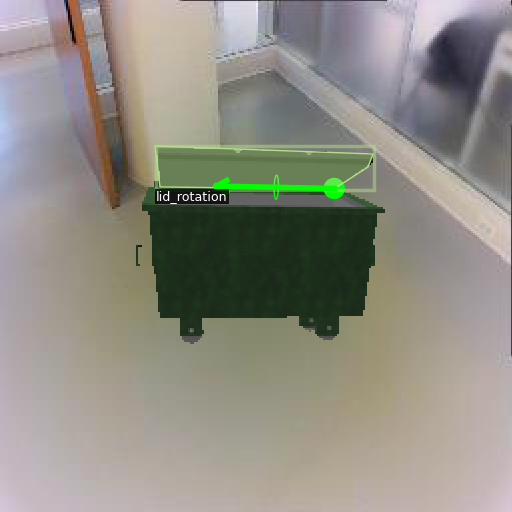} \\

\small{\opdrcnnop} &
\imgclip{0}{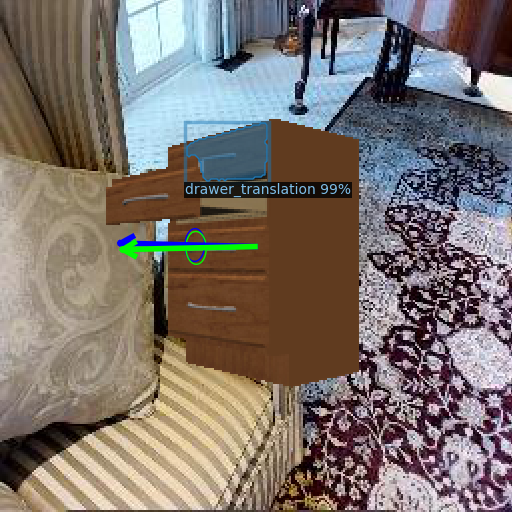} &
\imgclip{0}{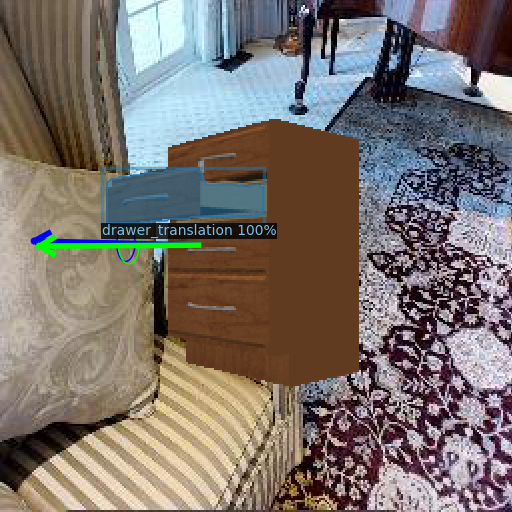} &
\imgclip{0}{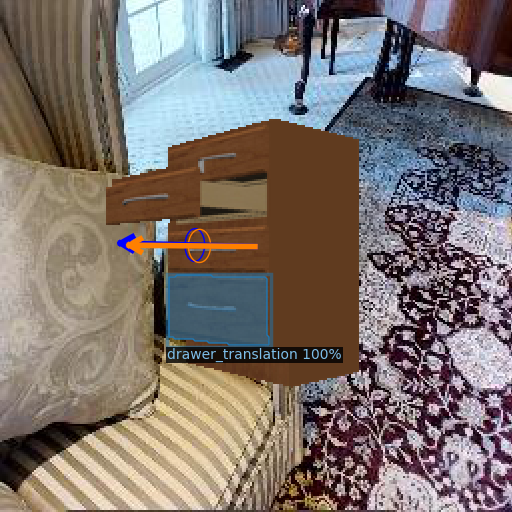} &
\imgclip{0}{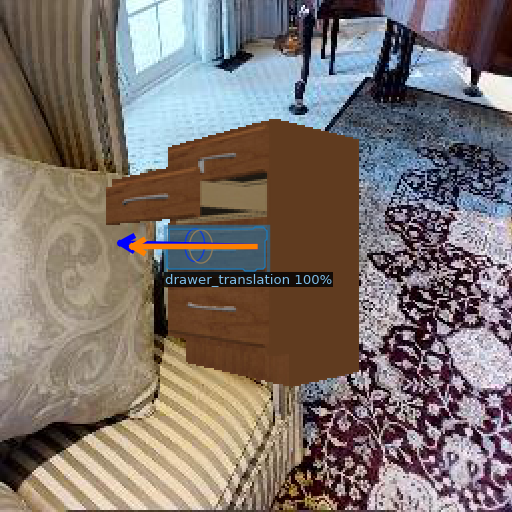} &
\imgclip{0}{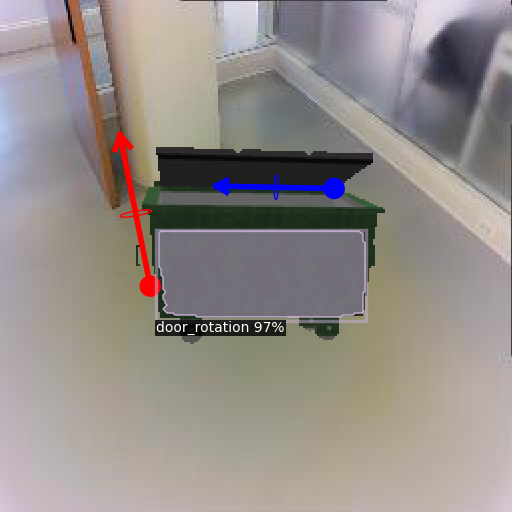} \\
\small{Axis (origin) error} & 2.446 & 3.721 & 5.127 & 6.669 & 79.841 (0.464)\\

\small{\opdformerp} &
\imgclip{0}{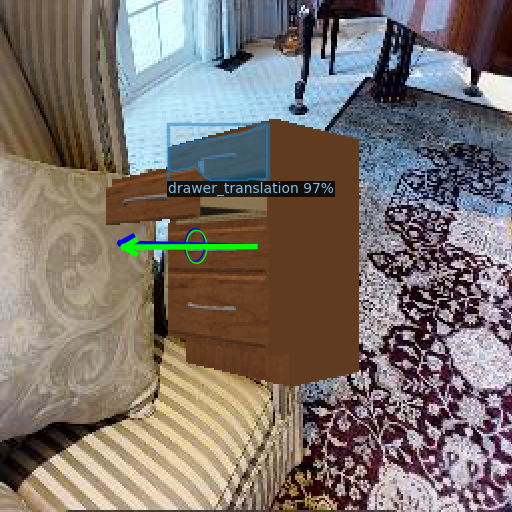} & 
\imgclip{0}{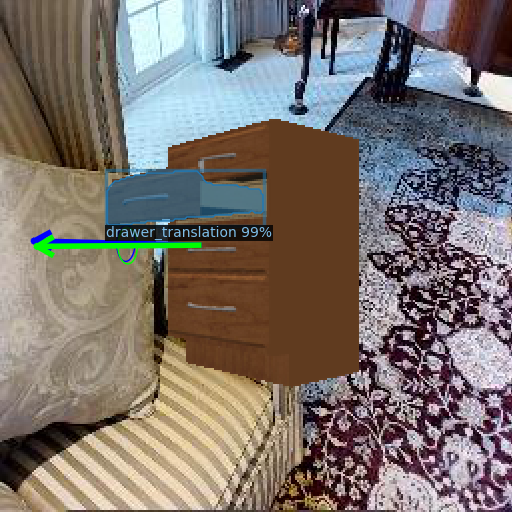} &
\imgclip{0}{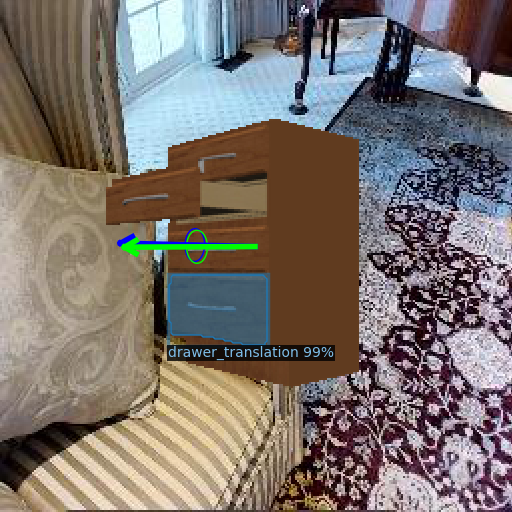} &
\imgclip{0}{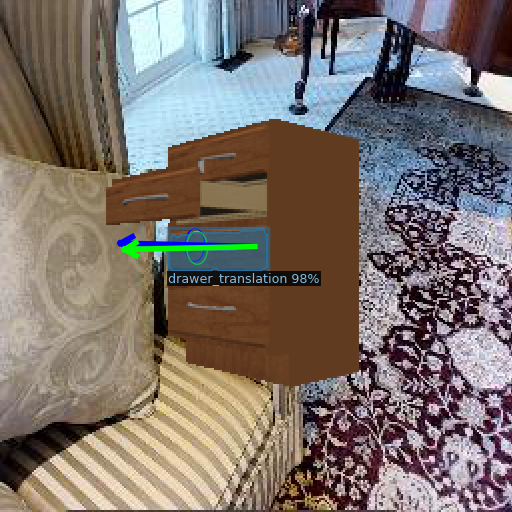} &
\imgclip{0}{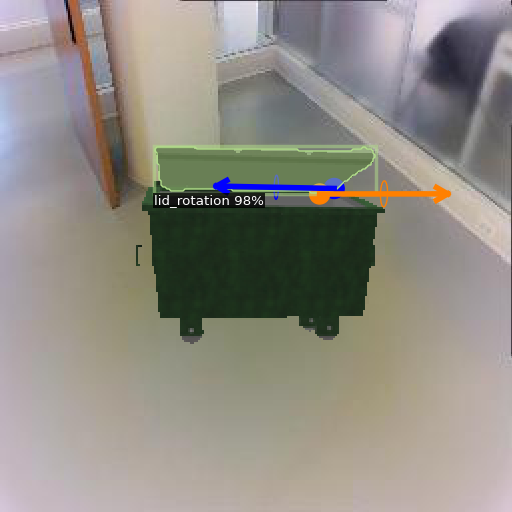}\\
\small{Axis (origin) error} & 2.368 & 2.517 & 2.768 & 3.170 & 5.939 (0.179)\\

\midrule
\textbf{\ourdatareal} \\
\midrule
\small{GT} &
\imgclip{0}{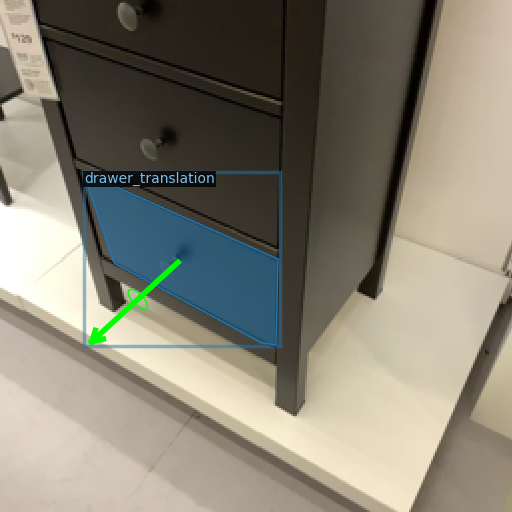} & 
\imgclip{0}{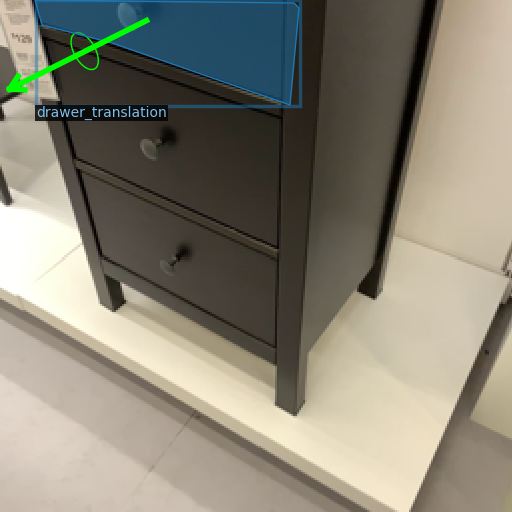} &
\imgclip{0}{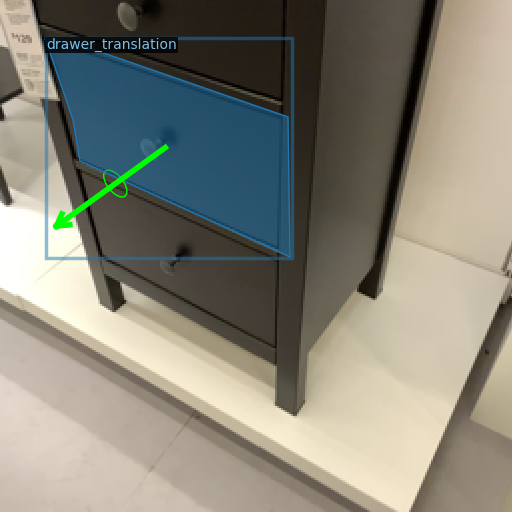} &
\imgclip{0}{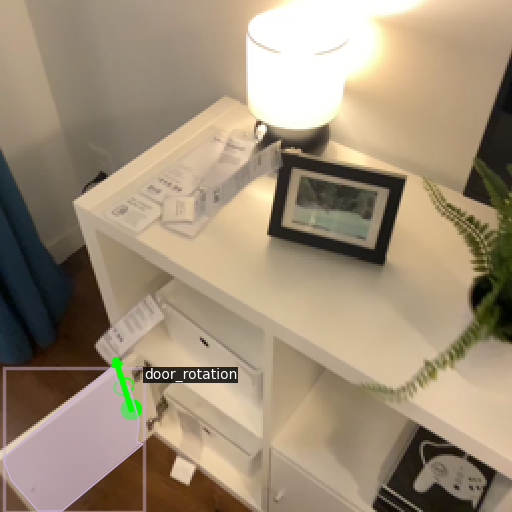} &
\imgclip{0}{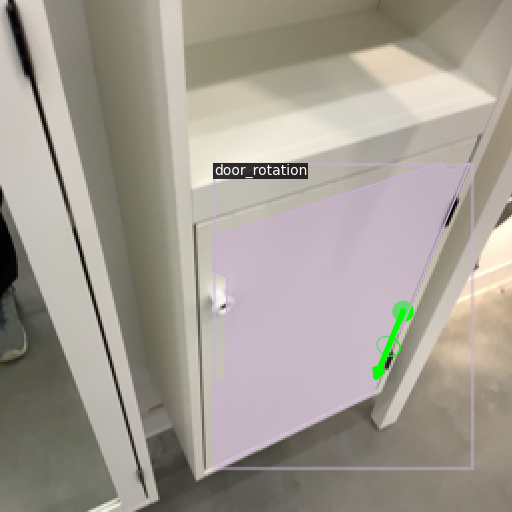} \\

\small{\opdrcnnop} &
\imgclip{0}{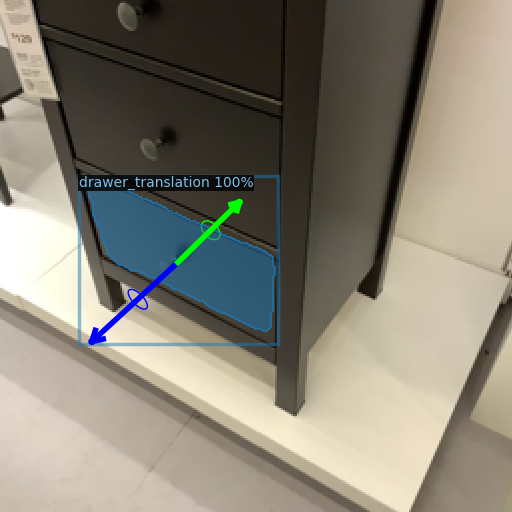} & 
\imgclip{0}{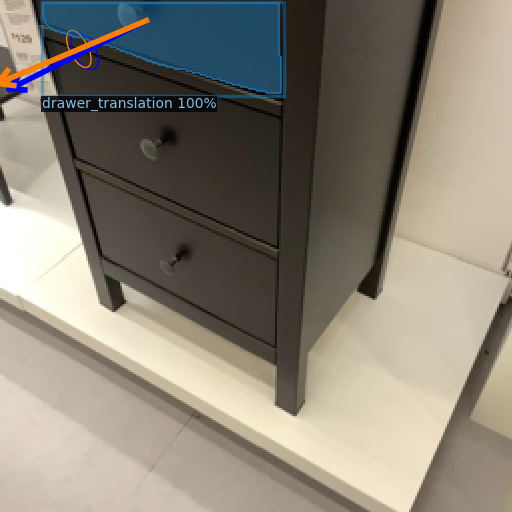} &
\imgclip{0}{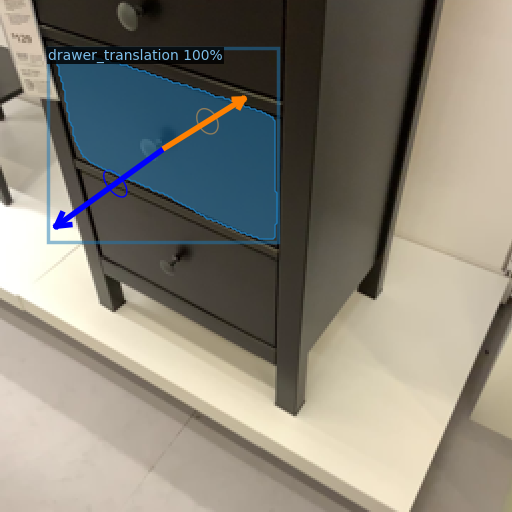} &
\imgclip{0}{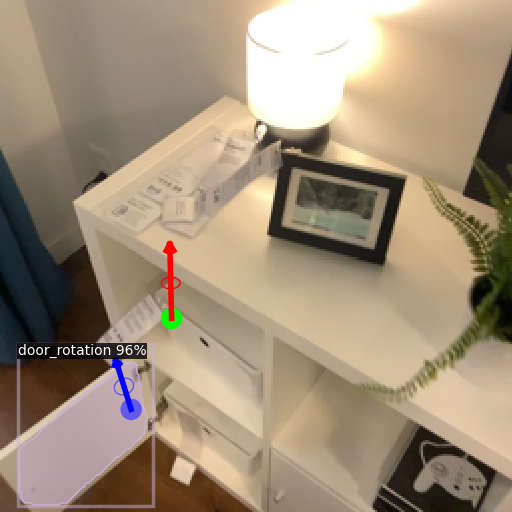} &
\imgclip{0}{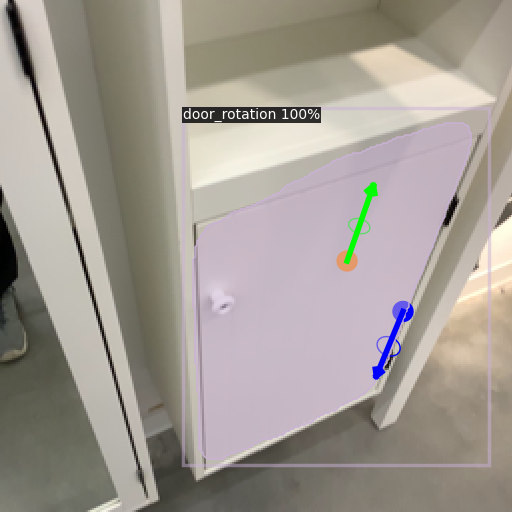} \\
\small{Axis (origin) error} & 4.753 & 5.492 & 5.262 & 12.491 (0.093) & 1.085 (0.047) \\

\small{\opdformerp} &
\imgclip{0}{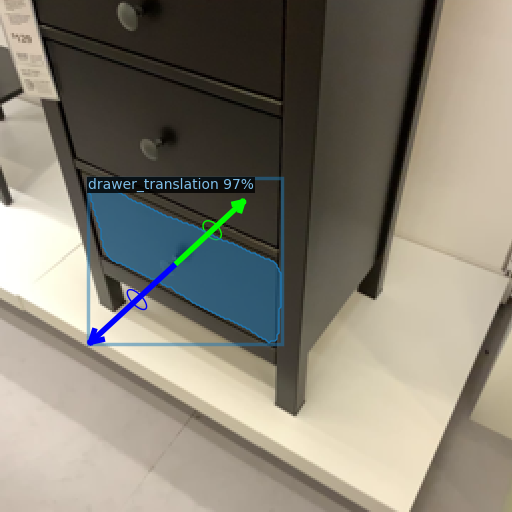} & 
\imgclip{0}{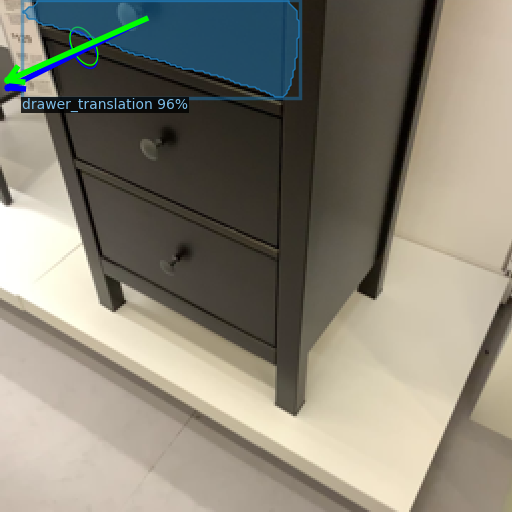} &
\imgclip{0}{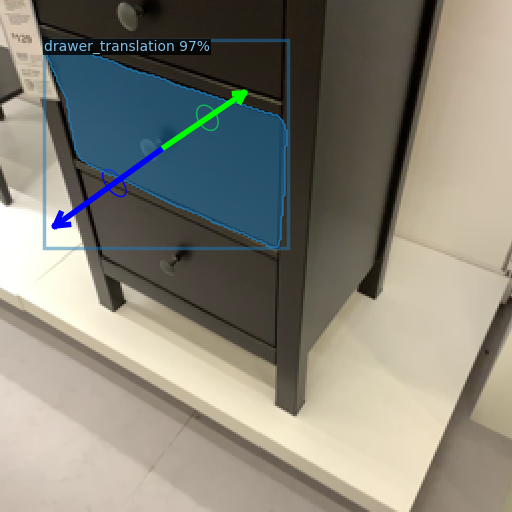} &
\imgclip{0}{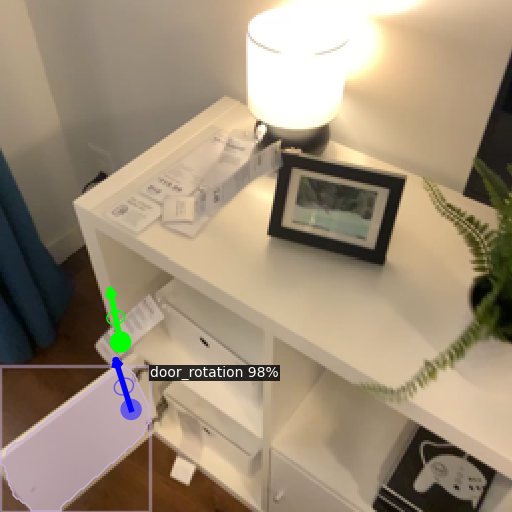} &
\imgclip{0}{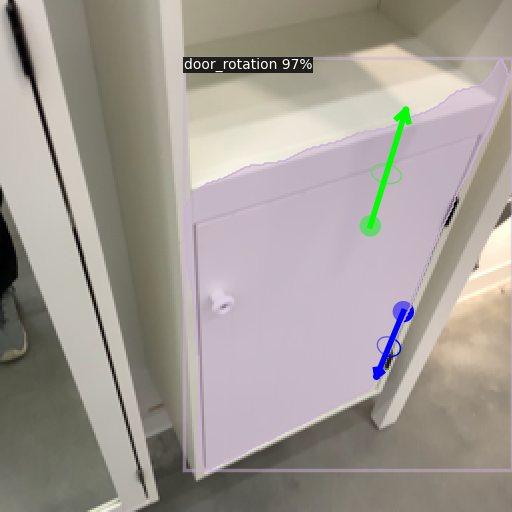} \\
\small{Axis (origin) error} & 2.409 & 2.712 & 2.690 & 2.965 (0.099) & 3.045 (0.113)\\

\bottomrule
\end{tabularx}
\caption{Qualitative results from the \ourdatacad and \ourdatareal~\cite{jiang2022opd} val sets.
The first row in each group of five rows is the ground truth (GT) with the motion axis shown in green.
The following rows show the results of \opdrcnnop and \opdformerp and the axis error (with origin error in parenthesis if the motion type is rotation).
If the predicted axis is close to the GT (within $5^\circ$), the predicted axis is shown in \textcolor{green}{green}. The predicted axis color is \textcolor{orange}{orange} if the angle difference is between $5^\circ$ and $10^\circ$, and \textcolor{red}{red} if the angle difference is greater than $10^\circ$.
If the motion type is rotation, the axis origin is visualized following the same color setting as the axis with the thresholds 0.1 and 0.25.
For examples from both datasets, we see that \opdformerp performs robustly in the prediction of motion axis and motion origin.
}
\label{fig:vis-compare-real-synth-single}
\end{figure*}

We provide additional qualitative results on both the \ourdatacad and \ourdatareal datasets.
\Cref{fig:vis-compare-real-synth-single} shows the qualitative results for selected models on the \ourdatacad and \ourdatareal datasets.
Overall, we again see that our proposed \opdformerbaseline variants have better performance than \opdrcnn methods and can predict more consistent motion parameters.

\begin{figure*}
\centering
\setkeys{Gin}{width=\linewidth}
\begin{tabularx}{\textwidth}{@{}Y Y Y Y Y Y Y@{}}
\toprule


\small{GT} &
\imgclip{0}{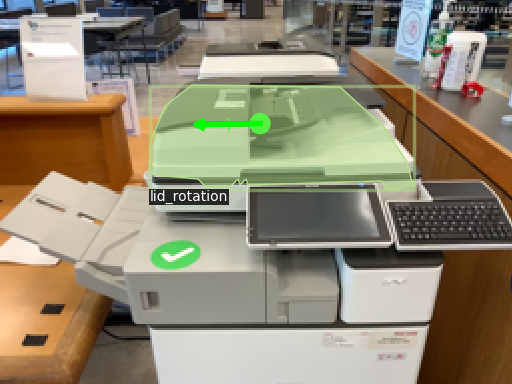} &
\imgclip{0}{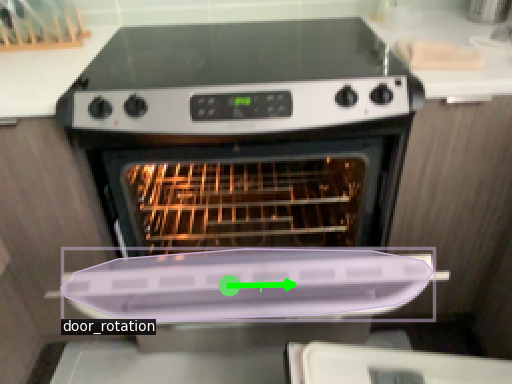} &
\imgclip{0}{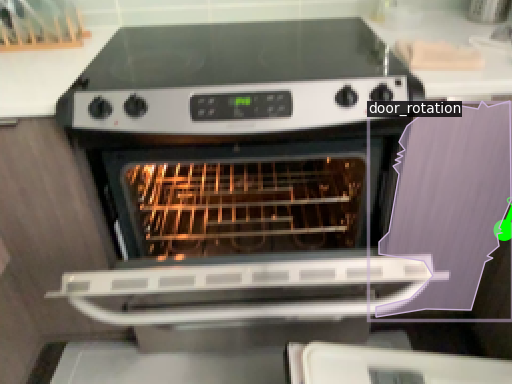} &
\imgclip{0}{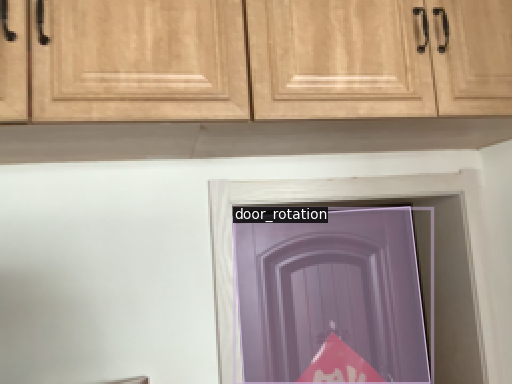} &
\imgclip{0}{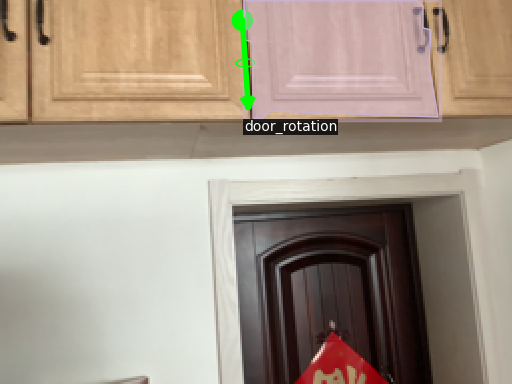} &
\imgclip{0}{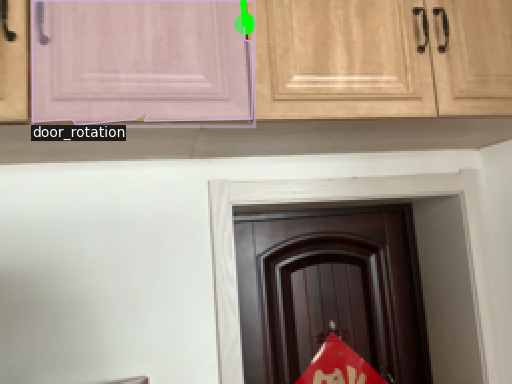} \\

\small{\opdformerc} &
\imgclip{0}{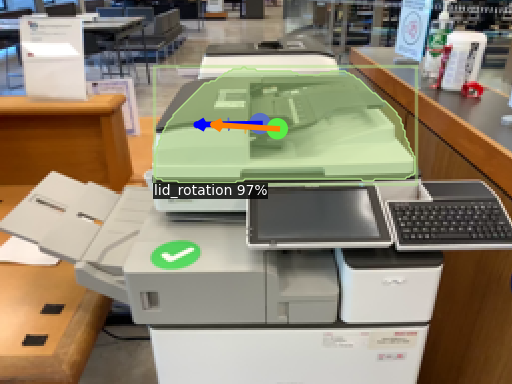} &
\imgclip{0}{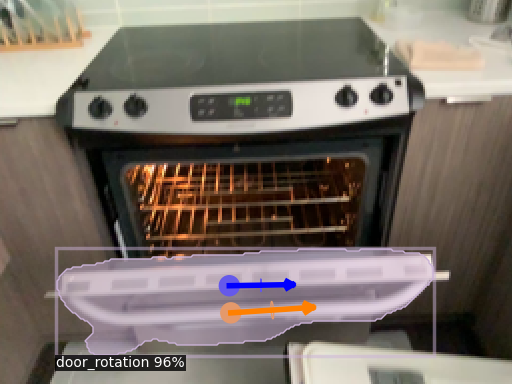} &
Miss &
Miss &
\imgclip{0}{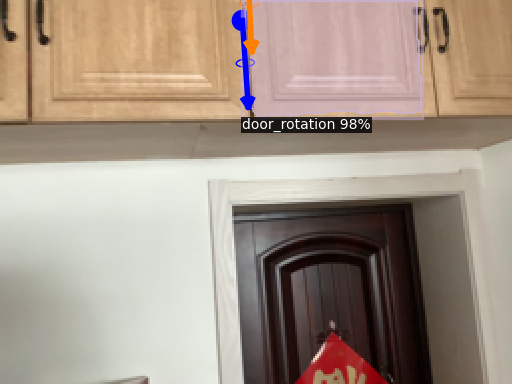} &
\imgclip{0}{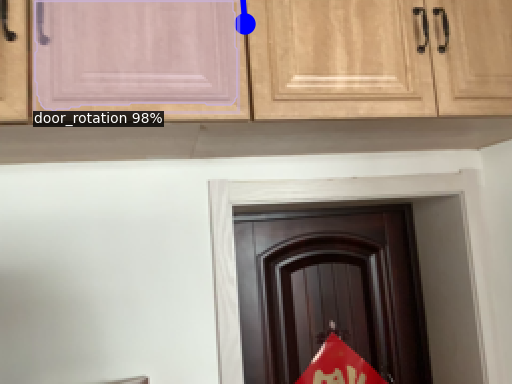} \\
\small{Axis (origin) error} & 5.326 (0.014) & 7.127 (0.179) & - & - & 9.593 (0.014) & 10.547 (0.016)\\

\small{\opdformero} &
\imgclip{0}{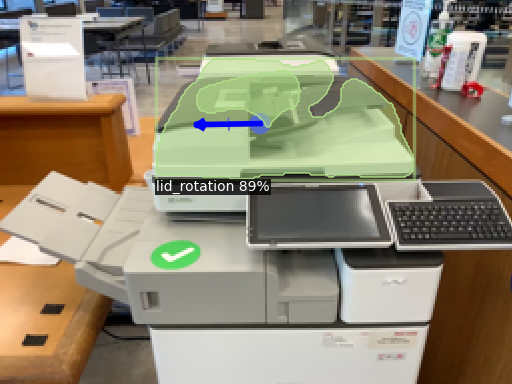} &
\imgclip{0}{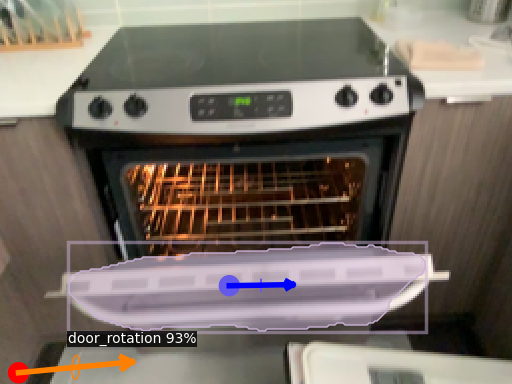} &
Miss &
\imgclip{0}{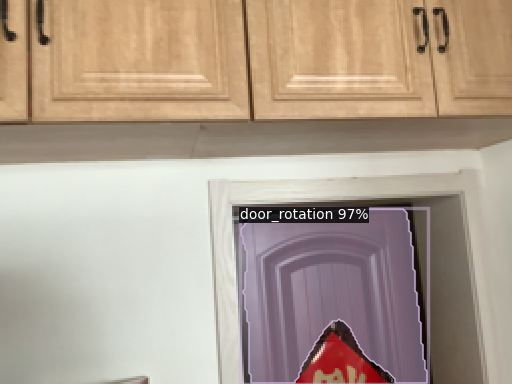} &
Miss &
Miss \\
\small{Axis (origin) error} & 12.476 (0.484) & 5.317 (0.317) & - &3.948 (0.217) & - & - \\

\small{\opdformerp} &
\imgclip{0}{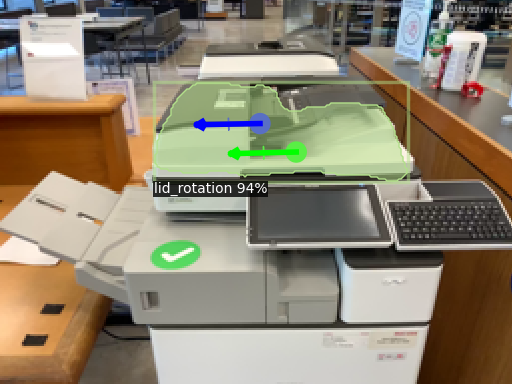} &
\imgclip{0}{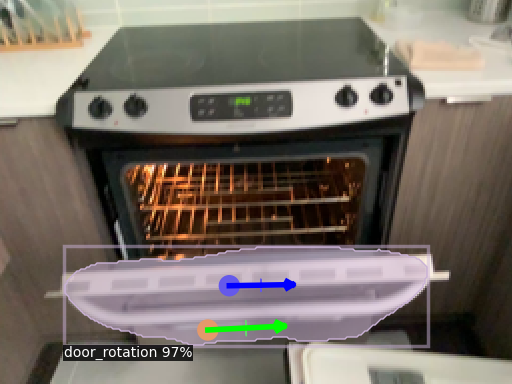} &
\imgclip{0}{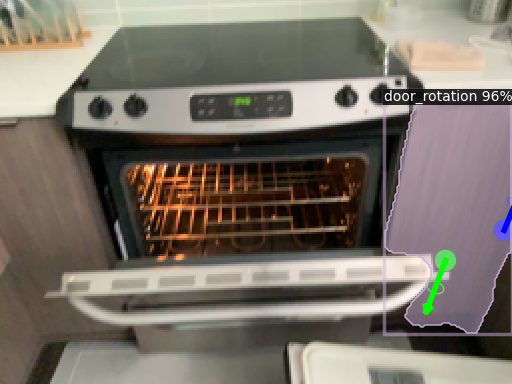} &
\imgclip{0}{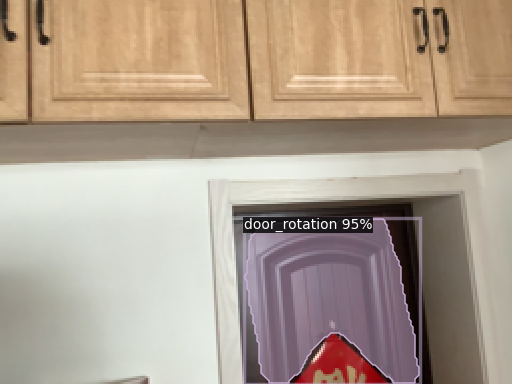} &
\imgclip{0}{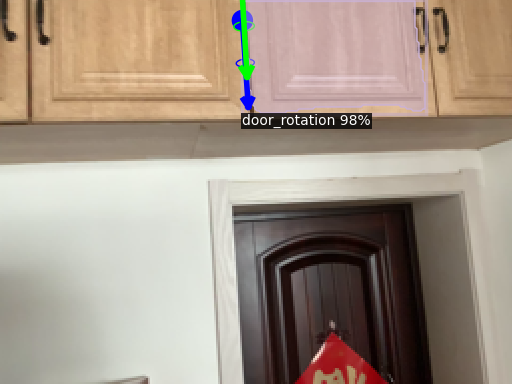} &
\imgclip{0}{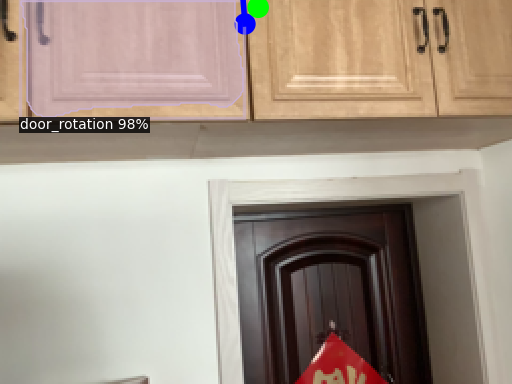} \\
\small{Axis (origin) error} & 4.425 (0.061) & 2.756 (0.129) & 2.895 (0.089) & 4.229 (0.334) & 3.624 (0.025) & 4.827 (0.010)\\

\bottomrule
\end{tabularx}
\caption{ 
Qualitative results for different \opdformerbaseline variants on \ourdatamulti. If the part is not predicted, we show ``Miss''.
If the predicted axis is close to the GT (within $5^\circ$), the predicted axis is shown in \textcolor{green}{green}. The predicted axis color is \textcolor{orange}{orange} if the angle difference is between $5^\circ$ and $10^\circ$, and \textcolor{red}{red} if the angle difference is greater than $10^\circ$.
If the motion type is rotation, the axis origin is visualized following the same color setting as the axis with the thresholds 0.1 and 0.25.
The first column shows that \opdformerp has more accurate motion prediction for the \lid part category.
From the other columns, we see that \opdformerp has less missed detections, and can performs better when there are multiple openable objects.
From the first two columns, we see that \opdformerp can outperform other variants in motion prediction even if the detection is similar.}
\label{fig:vis-compare-opdformer}
\end{figure*}

We also present qualitative visualizations of how the different variants of \opdformerbaseline performs on \ourdatamulti dataset (see   \Cref{fig:vis-compare-opdformer}). 
From the first two columns, we see that \opdformerp can perform better even when the detected mask is similar to other variants.
Overall, \opdformerp has better prediction ability when there are multiple object present, illustrating the importance of predicting the object pose on a per-part basis.

\subsection{Visualizing the transformer attention masks}
\label{sec:supp-attention}

\begin{figure*}
\centering
\setkeys{Gin}{width=\linewidth}
\begin{tabularx}{\textwidth}{Y Y Y Y Y Y Y}

\toprule

\small{GT} &
\imgclip{0}{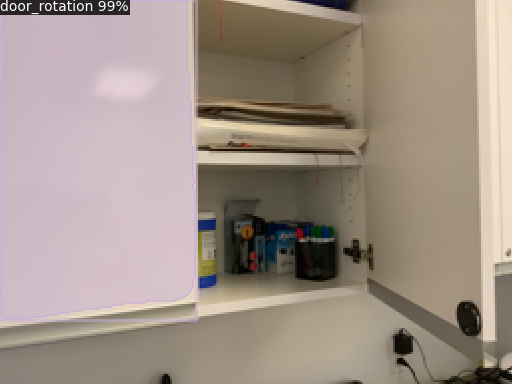} & 
\imgclip{0}{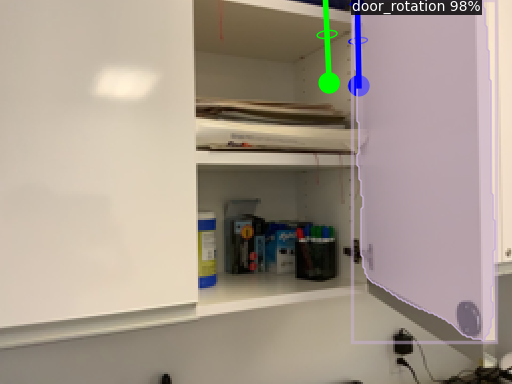} &
\imgclip{0}{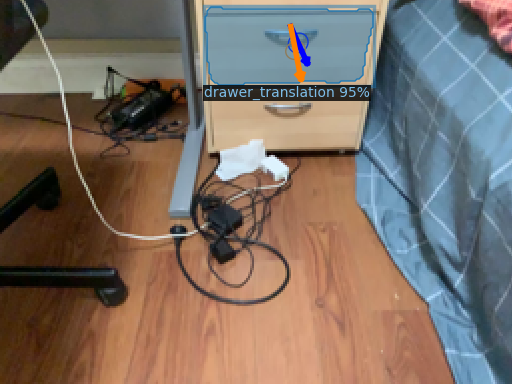} &
\imgclip{0}{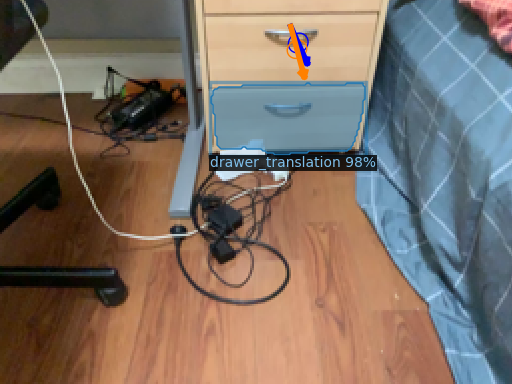} &
\imgclip{0}{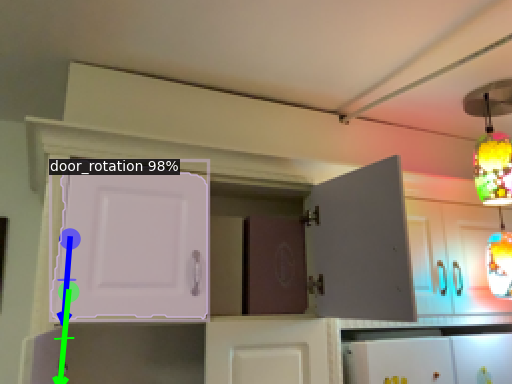} &
\imgclip{0}{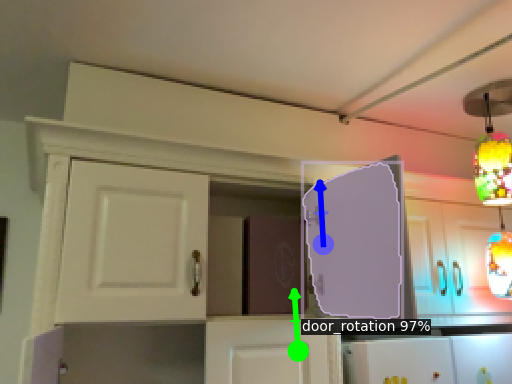}\\

\small{$1/32$} &
\imgclip{0}{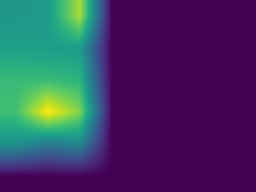} & 
\imgclip{0}{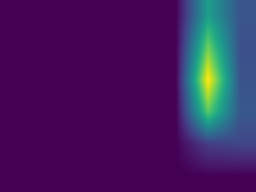} &
\imgclip{0}{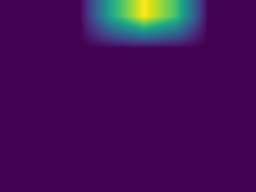} &
\imgclip{0}{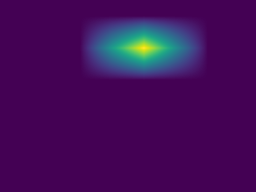} &
\imgclip{0}{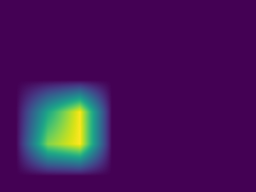} &
\imgclip{0}{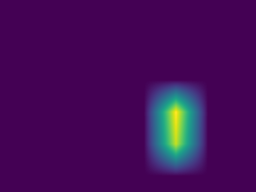}\\

\small{$1/16$} &
\imgclip{0}{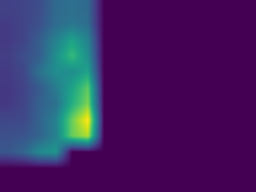} & 
\imgclip{0}{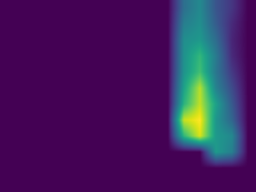} &
\imgclip{0}{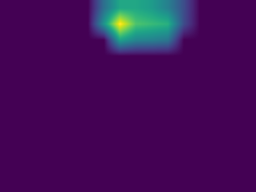} &
\imgclip{0}{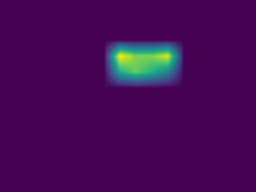} &
\imgclip{0}{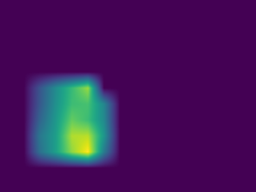} &
\imgclip{0}{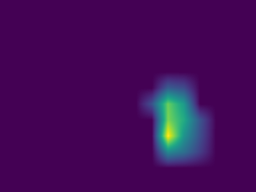}\\

\small{$1/8$} &
\imgclip{0}{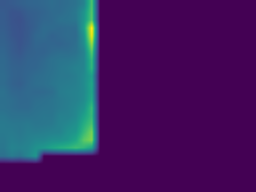} & 
\imgclip{0}{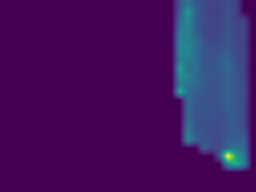} &
\imgclip{0}{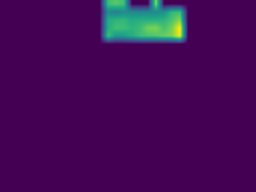} &
\imgclip{0}{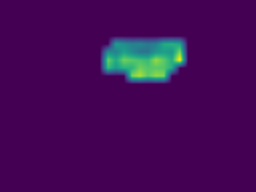} &
\imgclip{0}{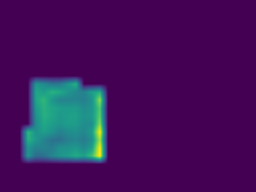} &
\imgclip{0}{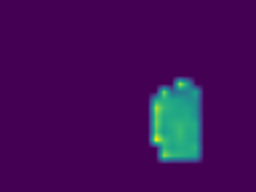}\\

\bottomrule
\end{tabularx}
\caption{
Visualization of masked attention for different image feature resolutions (from top to bottom, 1/32, 1/16 and 1/8 resolution).
We see high weights assigned to the detected part regions, and especially close to edges and corners of the parts.
}
\label{fig:vis-attention}
\end{figure*}

We visualize the attention maps of the transformer architecture following the same layer selection for the masked attention as Mask2Former~\cite{cheng2021masked}.
Figure \ref{fig:vis-attention} shows the visualizations.
We choose the attention maps of the last three masked attention layers, which use image features with different resolutions.
From the visualization, we see that the masked attention assigns high weight on the openable parts.

\subsection{Analysis of object part consistency}
\label{sec:supp-part-consistency}

As noted in the main paper, the motion parameters are highly correlated within an object, as well as across objects.
In a single object, parts with the same part category are likely to have similar motion types (e.g., all doors in one cabinet are either all rotating or sliding).  Similarly, the motion axes tend to be consistent across the parts (e.g. all drawers for a cabinet will tend to translate in the same direction, while a cabinet with both drawers and door will tend to have rotation axis for the doors that are perpendicular to the drawers' translation axes). Thus, the motion axes of different parts are likely parallel or perpendicular to each other.  Rotational motion origins follow a similar layout for rotating parts (e.g., the edge of rotating parts is constrained by the part position).

We investigate the consistency of the ground truth (GT) motions in the datasets, as well as how consistent the predictions from our model variants are.  We show that our \opdformerp provides the most consistent predictions.

\begin{table}[t]
\centering
\resizebox{\linewidth}{!}
{
\begin{tabular}{@{} ll rrr r  @{}}
\toprule 
      &       & \multicolumn{3}{c}{axis $\uparrow$} & type $\uparrow$ \\
\cmidrule(l{0pt}r{2pt}){3-5} \cmidrule(l{2pt}r{0pt}){6-6}
Dataset & Model & $1^\circ$ & $5^\circ$ & $10^\circ$ & \\
\midrule
\multirow{4}{*}{\ourdatacad} 
& \opdrcnnc~\cite{jiang2022opd} & 0.05 & 0.47 & 0.75 & 0.99 \\
& \opdrcnnop & 0.15 & 0.75 & 0.89 & 0.99 \\
& \opdformerc & 0.46 & 0.87 & \best{0.92} & 0.99  \\
& \opdformerp &  \best{0.72} & \best{0.89} & \best{0.92} & 0.99 \\
\midrule
\multirow{4}{*}{\ourdatareal} 
& \opdrcnnc & 0.02 & 0.20 & 0.43  & 0.95 \\
& \opdrcnnop & 0.03 & 0.38 & 0.68 & 0.93\\
& \opdformerc & 0.15 & 0.78 & \best{0.89} & 0.99 \\
& \opdformerp & \best{0.35} & \best{0.82} & \best{0.89} & 0.98 \\
\midrule
\multirow{4}{*}{\ourdatamulti} 
& \opdrcnnc & 0.02 & 0.29 & 0.57 & 0.97 \\
& \opdrcnnop & 0.02 & 0.28 & 0.56 & 0.96 \\
& \opdformerc & 0.18 & 0.79 & 0.88 & 0.98\\
& \opdformerp & \best{0.33} & \best{0.84} & \best{0.90} & 0.98 \\
\bottomrule
\end{tabular}
}
\caption{
Consistency of motion type and motion axis predictions on the val set of the three datasets.
For each part pair we check whether motion axes are parallel or perpendicular within three thresholds ($1^\circ$, $5^\circ$, and $10^\circ$).
For each part pair of the same category we check whether predicted motion types are the same.
We report averaged scores over all valid images.
For experiments in \ourtask we evaluate the consistency for each object in the image instead of the whole image.
}
\label{tab:results-consistency}
\end{table}

\begin{figure}
\centering
\setkeys{Gin}{width=\linewidth}
\begin{tabularx}{\linewidth}{Y Y}
\imgclip{0}{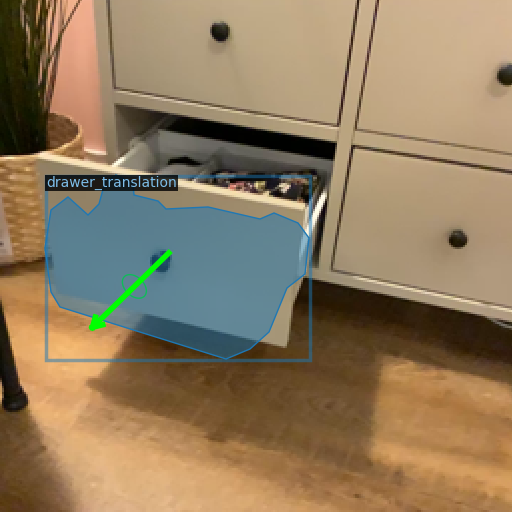} &
\imgclip{0}{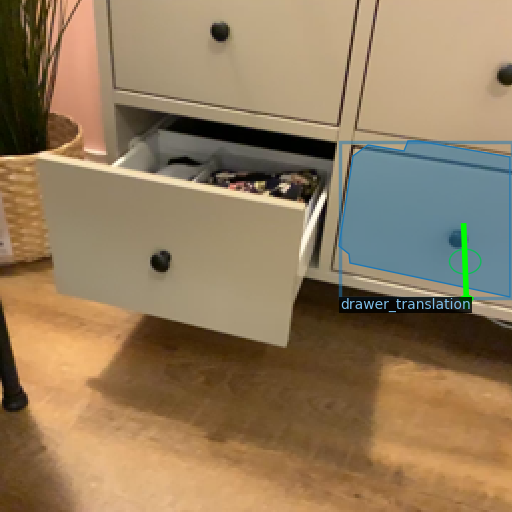} \\
\end{tabularx}
\caption{Example of inaccurate axis annotations in \ourdatareal from \citet{jiang2022opd}.
The parallel motion axes of the two drawers in the same cabinet have about 6.29 degree error. }
\label{fig:vis-bad-gt}
\end{figure}

\mypara{GT Motion Consistency.}
We check how \emph{consistent} the motion types and motion axes are across the single object datasets, \ourdatacad and \ourdatareal, by measuring the percentage of part pairs in the same object that: 1) have the same motion type given the same part category; and 2) are parallel or perpendicular to each other.
For motion type, all part pairs in both datasets are consistent according to our observation.
For the motion axes, we measure the axis consistency for three different angle thresholds (1$^\circ$, 5$^\circ$, and 10$^\circ$).
\ourdatacad matches our hypothesis for all objects across all thresholds, while in \ourdatareal all pairs matched at $10^\circ$, 97\% matched at $5^\circ$, and 68\% pairs at $1^\circ$.
Upon inspection, many non-consistent cases are due to small inaccuracies in the ground truth annotations for the motion axis in \ourdatareal.
This inconsistency is likely due to noise in the annotation.
\Cref{fig:vis-bad-gt} shows an example where there is a slight inconsistency.
The motion axis should have the same direction between the two drawers, but the GT motion axis is slightly different with a 6.29-degree error.

\mypara{Prediction Consistency.}
To evaluate motion consistency for predicted joints, we also measure pair-wise consistency of part motion predictions.
\Cref{tab:results-consistency} shows the results for \ourdatacad, \ourdatareal and \ourdatamulti.
We see that \opdformerbaseline predictions are much more consistent than \opdrcnn.
This is likely due to the overall better pose prediction and the self-attention mechanism in \opdformerbaseline that allows it to better capture relationships with other parts.
In addition, almost all predictions on \ourdatacad are parallel or perpendicular within $5^\circ$ and $10^\circ$, where \opdformerp has the best consistency. For \ourdatareal and \ourdatamulti, the consistency is low at $1^\circ$ but fairly good at $5^\circ$ and $10^\circ$.

\end{document}